\documentclass[10pt,journal,cspaper,compsoc]{IEEEtran}

\ifCLASSOPTIONcompsoc
  \usepackage[nocompress]{cite}
\else
  \usepackage{cite}
\fi

\ifCLASSINFOpdf
   \usepackage[pdftex]{graphicx}
   \graphicspath{{../pdf/}{../jpeg/}}
   \DeclareGraphicsExtensions{.pdf,.jpeg,.png,.jpg}
\else
   \usepackage[dvips]{graphicx}
   \graphicspath{{../eps/}}
   \DeclareGraphicsExtensions{.eps}
\fi

\usepackage{hyperref}
\usepackage{amsmath}
\usepackage{algorithm}
\usepackage{algorithmic}
\usepackage{amsfonts}
\usepackage{wrapfig}
\usepackage{setspace}

\ifCLASSOPTIONcompsoc
  \usepackage[caption=false,font=footnotesize,labelfont=sf,textfont=sf]{subfig}
\else
  \usepackage[caption=false,font=footnotesize]{subfig}
\fi

\usepackage{stfloats}

\usepackage{titlesec}

\titlespacing*{\section}{0pt}{2ex plus 1ex minus 1ex}{1ex plus 1ex minus 1ex}
\titlespacing*{\subsection}{0pt}{1ex plus 0.5ex minus 0.5ex}{0.5ex plus 0.5ex minus 0.5ex}

\def\ie{{\em i.e.}}

\usepackage{ragged2e}
\usepackage{booktabs}
\usepackage{color}
\usepackage{multirow}
\usepackage{bm}
\usepackage{bbm}
\usepackage{balance}

\newcommand{\AlAg}[1]{{#1}}

\begin{document}

\title{TexHOI: Reconstructing Textures of 3D Unknown Objects in Monocular Hand-Object Interaction Scenes}


\author{Alakh~Aggarwal,~Ningna~Wang,~Xiaohu~Guo
\IEEEcompsocitemizethanks{\IEEEcompsocthanksitem A. Aggarwal, N. Wang and X. Guo are with the Department of Computer Science, The University of Texas at Dallas, Richardson, TX 75083.
\IEEEcompsocthanksitem Corresponding Author: X. Guo, Email: xguo@utdallas.edu
}}



\IEEEtitleabstractindextext{%
\begin{abstract}
        \justifying Reconstructing 3D models of dynamic, real-world objects with high-fidelity textures from monocular frame sequences has been a challenging problem in recent years. This difficulty stems from factors such as shadows, indirect illumination, and inaccurate object-pose estimations due to occluding hand-object interactions. To address these challenges, we propose a novel approach that predicts the hand’s impact on environmental visibility and indirect illumination on the object's surface albedo. Our method first learns the geometry and low-fidelity texture of the object, hand, and background through composite rendering of radiance fields. Simultaneously, we optimize the hand and object poses to achieve accurate object-pose estimations. We then refine physics-based rendering parameters—including roughness, specularity, albedo, hand visibility, skin color reflections, and environmental illumination—to produce precise albedo, and accurate hand illumination and shadow regions. 
      Our approach surpasses state-of-the-art methods in texture reconstruction and, to the best of our knowledge, is the first to account for hand-object interactions in object texture reconstruction. Please check our work at: \textbf{{\color{blue}https://alakhag.github.io/TexHOI-website/}} \\
      \textbf{This work has been submitted to the IEEE for possible publication. Copyright may be transferred without notice, after which this version may no longer be accessible.}
\end{abstract}

\begin{IEEEkeywords}
Radiance Field, Differentiable Renderer, Spherical Gaussian, Albedo Prediction
\end{IEEEkeywords}}

\maketitle

\IEEEdisplaynontitleabstractindextext

\IEEEpeerreviewmaketitle

\IEEEraisesectionheading{\section{Introduction}}
    \label{sec:intro}
    \IEEEPARstart{E}{very} day, we interact with common objects like sugar boxes, snack containers, and cleaning supplies. Our hands naturally pick them up, move them, and manipulate them to complete routine tasks. These seemingly simple interactions, however, involve complex hand-object dynamics that challenge even the most advanced computer vision systems in accurately modeling and replicating their texture in virtual environments.
    
    In recent years, there has been an increase in the availability of standardized datasets designed to capture hand-object interactions, such as HO3D~\cite{hampali2020honnotate} and DexYCB~\cite{chao2021dexycb}, which are built on YCB-models~\cite{xiang2017posecnn}, or Contact-Pose~\cite{brahmbhatt2020contactpose} dataset. These datasets provide crucial insights into how hands manipulate everyday objects, like sugar boxes and snack containers, in dynamic scenarios. However, a major challenge remains in effectively utilizing these datasets due to the complex shadowing and reflection effects that hands cast on object surfaces. Accurately capturing and rendering these effects is critical for a wide range of applications, from enhancing realism in virtual and augmented reality environments to improving robotic manipulation and object recognition systems.
    Several recent works have focused on reconstructing hand-object interactions. Methods like BundleSDF~\cite{wen2023bundlesdf}, DiffHOI~\cite{ye2023diffusion}, and HOLD~\cite{fan2024hold} primarily address geometry reconstruction. However, these methods struggle with generating accurate textures under varying environmental illumination conditions. Moreover, shadows and indirect illumination caused by hand movements introduce significant errors in texture prediction. To the best of our knowledge, no recent work explicitly addresses the influence of shadows cast by dynamic elements like hands on 3D object reconstruction, particularly texture reconstruction.
    Other works such as PhySG~\cite{zhang2021physg}, InvRender~\cite{zhang2022modeling}, and RefNeRF~\cite{refnerf} focus on modeling the influence of environmental illumination on object surfaces based on material properties, typically represented as Spherical Harmonics or Spherical Gaussian. However, these methods do not account for interacting non-rigid parts like hands, making them unable to handle hand-induced shadows and reflections. Additionally, they rely on precise camera pose estimations based on object-coordinate systems, which can introduce errors akin to motion blur when these assumptions are violated.
    Given the above challenges, our primary research question is: How can we achieve accurate texture and geometry predictions in dynamic hand-object interaction scenarios using monocular camera data?

    In this paper, we present a novel framework to accurately predict the object's true albedo, specular properties, and surface geometry, given a sequence of frames captured by a single camera showing a hand interacting with a rigid object.
    Our focus is on accounting for the indirect effects of hand movements, such as shadows, reflections, and visibility changes, across different frames. Ultimately, we aim to enhance the realism and accuracy of virtual object rendering in complex hand-object interaction scenarios.
    
    Our approach begins by optimizing both hand and object poses using a composite rendering technique that separately models the hand, object, and background. In this process, we simultaneously refine hand and object poses while learning neural radiance fields for each entity, ensuring accurate geometric and pose representations.
    Using these fine-tuned poses, we represent the environmental illumination with Spherical Gaussian (SG) representations. This allows us to model the interaction between the object's material properties — such as specularity and roughness — and the environment.
    To efficiently compute the shadows and reflections caused by interacting hands, we introduce a novel representation of the hand using 108 parameterizable spheres. For each surface point on the object, we calculate the visibility and indirect illumination caused by each sphere by measuring the fraction of its projection that occludes the SG on the object’s surface. This method avoids the need for time-consuming ray-tracing~\cite{zhang2021nerfactor} or learning pseudo-ground truth visibility using MLPs~\cite{zhang2022modeling}.
    We then combine all the elements in the physics-based rendering (PBR) framework to obtain the final object color, which is compared with the ground truth camera-captured images to optimize all those physics-based rendering parameters, i.e., roughness, specularity, albedo, hand visibility, skin color reflections, and environmental illumination. 
    
    Our approach surpasses state-of-the-art methods in texture reconstruction quality, for dynamic hand-object interaction scenarios.
    We summarize our major contributions as follows:
    \begin{itemize}
        \item To the best of our knowledge, this is the first method to account for dynamic hand-object interaction in the context of texture prediction, for 3D object reconstruction from monocular videos.
        \item We introduce a novel hand representation where the canonical hand is simplified as $108$ parameterizable spheres. Skinning weights are allocated to each sphere based on the distance to surface points, enabling accurate pose transformations. The radius of each sphere is recalculated dynamically according to the transformed surface points. These spheres enable efficient calculation of Spherical Gaussian occlusion by the hand, eliminating the need for time-consuming ray-tracing computations for occlusion computation.
        \item We develop a two-stage framework for 3D object reconstruction in hand-object interactions. The first stage uses compositional volumetric rendering to optimize object and hand poses, while the second stage uses Spherical Gaussian-based surface rendering to refine object texture by removing hand occlusions and lighting effects, ensuring high-fidelity reconstruction.
\end{itemize}

\section{Related Works}

    Creating realistic rendered images of a target object, for its re-usability in varying virtual environment has been a major focus in the study of computer graphics for much of its history. Inverse rendering of an object from their visual data refers to predicting camera poses and scene descriptions, including the geometry of the object, the reflectance of the object surface, which consists of its albedo texture and other microfacet surface properties like roughness, specularity, etc., and environment lighting used to illuminate the scene~\cite{marschner1998inverse}. Many research works have built the basic framework for upcoming advancements in this area of research.
    
    \subsection{Traditional Inverse Rendering for 3D Object Reconstruction}
        Traditional inverse rendering approaches broadly divided the research field into three categories - \textbf{inverse lighting} to predict the environment light that illuminates the scene in visual input such as images, \textbf{inverse reflectometry} to predict object texture or object BRDF from visual input, or \textbf{shape reconstruction} to obtain the geometry of the object. 
        Traditional approaches only focus on solving one of the three mentioned problems. They may solve inverse lighting problem~\cite{kawai1993radioptimization} which allows re-lighting of rendered scenes under different environment illumination~\cite{nimeroff1995efficient,teo1997cient}. Some other works solve inverse reflectometry problems to obtain realistic texture~\cite{turk1991generating,witkin1991reaction} or accurate reflective model~\cite{curless1996volumetric,eck1995multiresolution} on the object surface. Meanwhile, other approaches optimize object geometry~\cite{ozyecsil2017survey,su2014estimating,james2012straightforward,westoby2012structure}. These approaches assume other aspects of the inverse rendering, except their focus problem, to be known. To our knowledge, no traditional approach comprehensively addresses all aspects of the inverse rendering problem in a single study. Recent advancements in neural representations address the comprehensive computation of these different aspects.
    \subsection{Neural Representation in Inverse Rendering}
        In recent years, several recent neural network-based research works have sparked a wave of innovative methodologies to solve inverse rendering problems. These include volumetric rendering based approaches like NeRF~\cite{mildenhall2021nerf}, NeuS~\cite{wang2021neus}, DVGO~\cite{sun2022improved}, TensoRF~\cite{chen2022tensorf}, Instant-NGP~\cite{muller2022instant}, etc., surface rendering based approaches like IDR~\cite{yariv2020multiview} or point-based inverse rendering approaches like 3D-Gaussian-Splatting~\cite{kerbl20233d}.  Moreover, these approaches give a general framework solution for inverse rendering problems but do not solve complex problems that arise from varying material properties of objects, environment illumination in the scene, or occlusion and visibility handling in multi-object interactions in the scene. Applying the physics-based knowledge, and addressing the composition of multiple-object interactions in the scene further improves this research area, as will be seen in the next two subsections.
    \subsection{Physics-based Inverse Rendering}
        The Physics-based Rendering (PBR) equation gives the rendered output color $c$ at a surface point $x$ of an object along the view direction $\omega_o$, given the illumination on the surface $L(\omega_i,x)$ from incoming direction $\omega_i$, BRDF property of the object $\phi(x,\omega_o,\omega_i)$ and object normal $n$ at $x$. The equation is defined as follows:
        \begin{equation}
            \label{eq:pbr}
            c(\omega_o; x) = \int_{\omega_i \in \Omega^+(n)} L(\omega_i,x) \phi(x;\omega_o,\omega_i)(\omega_i \cdot n),
        \end{equation}
        where $\omega_i$ is a direction in hemi-sphere $\Omega^+(n)$, defined with the $up$ direction as surface normal $n$. 
        $L(\omega_i,x)$ is obtained by inverse lighting, $\phi(x;\omega_o,\omega_i)$ is obtained by inverse reflectometry and normal $n$ is obtained by shape reconstruction. Occlusions between the illumination source and surface point $x$ may break down $L(x)$ term, requiring visibility and indirect illumination calculation.
        Many recent approaches have done extensive research on improving the inverse rendering output by predicting the above terms of the PBR equation. NeRV~\cite{srinivasan2021nerv} assumes known illumination. Other methods like NeRD~\cite{boss2021nerd}, Neural Ray-Tracing~\cite{knodt2021neural}, NeRFactor~\cite{zhang2021nerfactor}, Neural-PIL~\cite{boss2021neural}, PS-NeRF~\cite{yang2022ps}, Neilf~\cite{yao2022neilf}, SAMURAI~\cite{boss2022samurai}, NVDiffRec~\cite{hasselgren2022shape}, Neural-PBIR~\cite{sun2023neural}, TensoIR~\cite{jin2023tensoir}, TensoSDF~\cite{li2024tensosdf}, MIRReS~\cite{dai2024mirres}, IRON~\cite{zhang2022iron} etc. use volumetric-rendering based methods to optimize neural radiance fields, and extract geometry, illumination, texture and reflectance model. NeRO~\cite{liu2023nero}, NeP~\cite{wang2024inverse}, PBIR-NIE~\cite{cai2024pbir} extend these methods to glossy or reflective surfaces.
        Relightable-3D-Gaussian~\cite{gao2023relightable} uses 3D Gaussian splats, and methods like IntrinsicAnything~\cite{chen2024intrinsicanything} learn diffusion priors to obtain accurate albedo and material properties. Methods like PhySG~\cite{zhang2021physg} and InvRender~\cite{zhang2022modeling} use surface-based rendering approaches and approximate illumination and BRDF as Spherical Gaussians (SGs). An SG comprises of a lobe with its direction $\xi$, intensity $\mu$ and width \AlAg{$\eta$}, and is defined on a spherical surface as follows:
        \begin{equation}
            \label{eq:sg}
            G(v; \xi,\mu,\AlAg{\eta}) = \mu e^{\AlAg{\eta} (v \cdot \xi - 1)},
        \end{equation}
        where, $v$ is any direction on the sphere and $G(v)$ is the value on the sphere along direction $v$.
        The above methods either ignore visibility computation~\cite{zhang2021physg}, perform computationally-extensive Monte-Carlo ray tracing~\cite{zhang2021nerfactor}, or estimate pseudo-ground truth visibility using a Multi-layer Perceptron (MLP)~\cite{zhang2022modeling}. Monte-Carlo ray-tracing can take up to ~30 minutes per frame on a single GPU to train. L-Tracing~\cite{chen2022tracing} utilizes sphere-tracing, and the concavity/convexity of geometry to avoid learning visibility. However, both InvRender~\cite{zhang2022modeling} and L-Tracing~\cite{chen2022tracing} work for a single static rigid object, and fail to handle multiple dynamic object visibility estimation. NeRF-DS~\cite{yan2023nerf} handles dynamic objects but does not utilize the PBR equation to obtain the reflectometry information in the scene. Instead, the method is only restricted to glossy objects.
        Moreover, all the above methods assume a single rigid object with accurate pose estimations in the scene. These methods do not handle multiple dynamic object interactions in the scene.
    \subsection{Handling Dynamic Elements in Multi-Object Interaction}
        To handle multiple object interactions, along with background, many recent researches have come up, with compositional-NeRF methods~\cite{wang2023learning,driess2023learning,wu2022object,yang2021learning}. HandNeRF~\cite{guo2023handnerf} extends this compositional-NeRF methodology for multiple interacting hands. Other methods  named HandNeRF~\cite{choi2024handnerf}, BundleSDF~\cite{wen2023bundlesdf}, HOLD~\cite{fan2024hold}, etc. focus on dynamic hand-object reconstruction. However, their primary focus is geometry reconstruction, and do not focus on high-fidelity texture reconstruction. Hence, they do not focus on physics-based accuracy in results.
        Our research primarily focuses on 3D geometry and high-fidelity texture reconstruction of hand-object interactions from single-camera data capture sequences. We aim to improve the output by resolving the baking of hand effects into the rendered object texture, and obtain a high-fidelity object albedo prediction following a two-stage training process. In the first stage, we fine-tune hand and object poses using a compositional volumetric rendering approach~\cite{fan2024hold}. In the second stage, we use a surface rendering approach with Spherical Gaussians to obtain different elements in the PBR equation, including hand effects on the object surface. In Sec.~\ref{sec:method}, we will describe the above two stages of training in-depth.

\section{Preliminaries}
    \subsection{Volumetric Rendering}
    \label{vol-rend}
           For each pixel in the image space, the volumetric rendering samples the points uniformly along a ray originating from the camera pointing in the direction of the pixel. 
           These points are distributed between the near and far bounds of the camera's viewing range. Based on the predicted density values at these uniformly sampled points, an importance sampling technique refines the sampling process, drawing points closer to the estimated implicit surface. For each sampled point, the color is predicted using the view direction and the point’s properties. 
            The predicted colors are then aggregated along the ray through an integral, resulting in the final color for that pixel. Formally, along a camera ray $r(t) = o + td$ with camera location $o$ and view direction $d$, near and far bounds $t_n$ and $t_f$, density at a point is estimated as $\sigma(r(t))$ and color at a point is estimated as $c(r(t), d)$. The expected final color on the pixel $C(r(t))$ can thus be expressed as the following integral:
            \begin{equation}
                C(r(t)) = \int_{t_n}^{t_f} T(t)\sigma(r(t))c(r(t),d)dt,
            \end{equation}
            where $T(t) = exp(-\int_{t_n}^t \sigma(r(s))ds )$ represents the accumulated transmittance, which accounts for the occlusion of light as it passes through the scene. This approach captures both the density and color contributions along the entire ray, resulting in realistic volumetric effects. Methods like NeRF~\cite{mildenhall2021nerf} use volumetric rendering.
            
    \subsection{Surface Rendering}
            The surface rendering focuses on identifying the exact intersection point $\hat{t}$ of the ray $r(\hat{t})=o+\hat{t}d$ with the implicit surface geometry using the sphere tracing approach. Once the intersection point is determined, the color at that point is predicted as $c(r(\hat{t}), d)$. After calculating the surface color, it is assigned to the corresponding pixel in the image, forming the final rendered image. Methods like IDR~\cite{yariv2020multiview}, PhySG~\cite{zhang2021physg}, etc. use surface rendering.

\begin{figure*}
            \centering
            \includegraphics[width=0.89\textwidth]{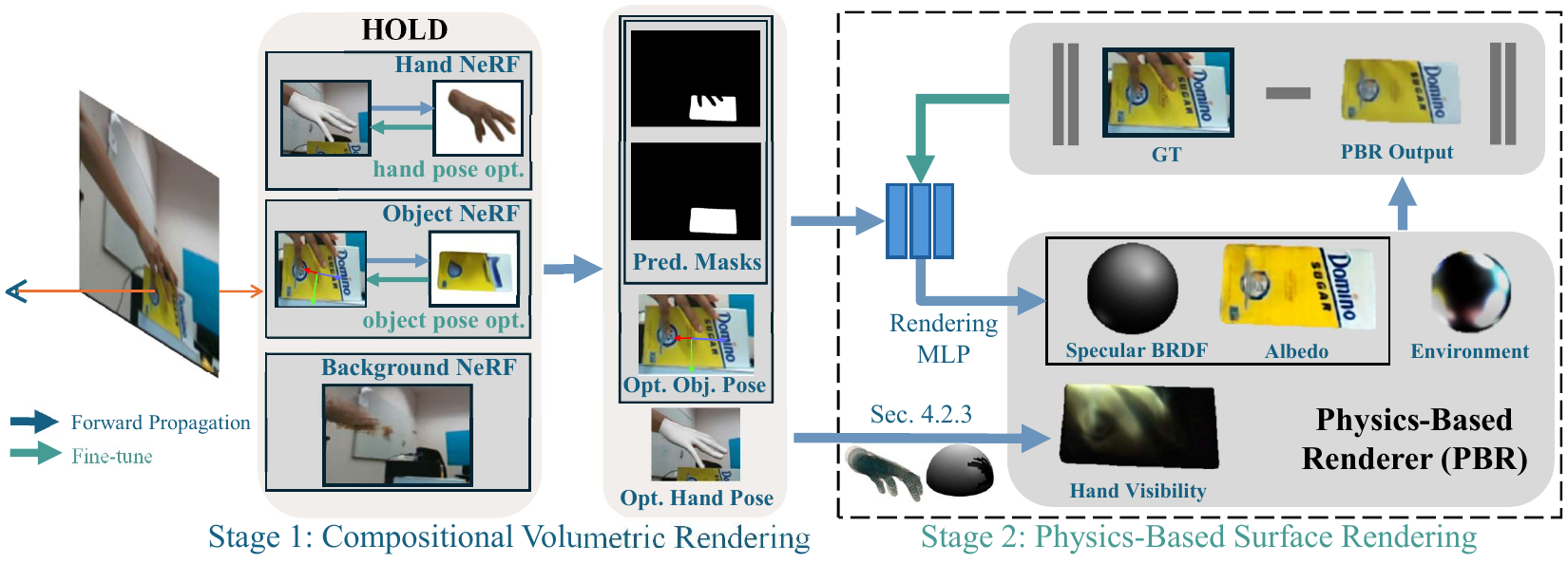}
            \caption{The overall pipeline of our proposed TexHOI method. In the first stage (Sec.~\ref{sec:stage1}), the hand and object poses are fine-tuned along with composite radiance fields of hand, object, and background. Using the predicted object segmentation and object geometry mask, in the second stage (Sec.~\ref{sec:stage2}), the optimized hand and object poses are used to accurately learn material properties on the object surface, \ie albedo, BRDF, and hand occlusions, using physics-based rendering with Spherical Gaussian approximations.}
            \label{fig:pipeline}
        \end{figure*}

\section{Methodology}\label{sec:method}
    This section outlines the methodological approach adopted to achieve high-fidelity geometry and texture reconstruction of hand-object interaction from monocular camera data sequence. We discuss in-depth both the stages of the two-stage process, called TexHOI, namely volumetric rendering based compositional NeRF as \textbf{Stage~1} (Sec.~\ref{sec:stage1}) to fine-tune object and hand pose and obtain a low-level of geometry and texture information, and surface rendering based Physics-Based Rendering (PBR) equation, approximated with Spherical Gaussians as \textbf{Stage~2} (Sec.~\ref{sec:stage2}) to generate hand shadows, reflections and environmental illumination, along with studying the material properties. We also discuss the data acquisition/processing steps in both stages, including hand and object pose estimation in stage 1, and visibility and reflectance property computation in stage 2. This two-stage process of computation of different elements leads to a high-fidelity albedo texture prediction. Fig.~\ref{fig:pipeline} shows the full pipeline.
    
    \subsection{Stage 1: Compositional Neural Radiance Field for Pose Refinement}
    \label{sec:stage1}
        The input of stage 1 of our system is the frame sequences from a monocular camera, along with an estimated hand and object poses and generates low-level geometry and texture predictions of hand and object. 
        
        This stage utilizes three Neural Radiance Fields (NeRF). NeRF for hand optimizes the hand pose transformation and generates hand from monocular camera frames in MANO~\cite{romero2022embodied} canonical pose. Similarly, a NeRF for the object optimizes the object pose transformation and generates the rigid object in its canonical pose, or world origin. Meanwhile, another NeRF learns background image representation. Finally, the outputs of the three NeRFs are composed to generate a composite rendered image similar to the ground-truth image frame. In this section, we will discuss in detail the above-mentioned NeRFs and their composition.
        
        The geometry prediction implicit network, along with the fine-tuned hand and object poses prepare the input of the second stage of our pipeline. 
        \subsubsection{Object NeRF}
            \label{object_nerf}
            Object NeRF uses object pose estimation to calculate neural network-based radiance field. The points sampled along the camera ray are \AlAg{inverse-transformed} to the object coordinate system using object pose. The density and color are thus predicted in the object coordinate system, or object canonical space.
            Given object pose with rotation $R$ and translation $t$, we get the canonical space point $x_{obj}$ from observation space point $x'$ as follows:
            \begin{equation}
                x_{obj} = R^{-1}(x' - t).
            \end{equation}

            Object NeRF, then, trains on the transformed points in object canonical space.
        \subsubsection{Hand NeRF}
            \label{hand_nerf}
            Calculating the radiance field for hand NeRF follows a similar setup as object NeRF. The points sampled along the camera ray are inverse-transformed to the hand's canonical pose, i.e., stretched hand, using MANO~\cite{romero2022embodied} pose parameters. Hence, we again need to perform the inverse transformation for hand.
            Given transformation matrices $T_i$ for $i^{th}$ of \AlAg{$n_{jnt}$} joints, with interpolated skinning weight $w_i(x')$ on the observation space point $x'$, its corresponding canonical space point $x_{hand}$ is calculated as follows:
            \begin{equation}
                x_{hand} = (\Sigma_{i=1}^{\AlAg{n_{jnt}}} (w_i(x') T_i) ^ {-1} x'.
            \end{equation}
        \subsubsection{Compositional NeRF}
            To generate a composite rendered image similar to a ground-truth frame, we also learn Background NeRF. After obtaining color and density predictions from each NeRF, we arrange all the sample points in sorted order from their distance. The other information at the sample points, i.e., density and color are also sorted similarly. We then calculate the aggregate integral and get the expected color at the pixel corresponding to the ray direction.
            
            Mathematically, sampling $n$ points for object and inverse transforming the points into object-coordinate system~\ref{object_nerf}, and $n$ points for hand and performing linear inverse blend-skinning into hand's canonical coordinate system~\ref{hand_nerf}, we obtain the foreground (hand and object) color $C_{fg}$ and foreground mask $M_{fg}$. The values are calculated by sorting the $2n$ points, and the predicted \AlAg{density} $\sigma_i$ and \AlAg{radiance $c_i$} values of each point. Similarly, using the calculated foreground mask $M_{fg}$ and the background color $C_{bg}$, we obtain final predicted color along a ray $r$ as follows:
            \begin{equation}
                C(r) = C_{fg}(r) + (1 - M_{fg}(r)) \cdot C_{bg}(r),
            \end{equation}
            where the foreground color is the volumetric rendering output of:
            \begin{equation}\label{eq:nerfcolor}
                C_{fg}(r) = \Sigma_{i=1}^{2n}\tau_i c_i.
            \end{equation}
            Here $\tau_i = exp(-\Sigma_{j<i}\sigma_j \delta_j) (1 - exp(-\sigma_i \delta_i))$ and $\delta_i$ is the distance between $i^{th}$ and $(i+1)^{th}$ sample points.
            Similarly, the foreground mask, is given as:
            \begin{equation}
                \label{eq:segmentation}
                \AlAg{M}_{fg}(r) = \Sigma_{i=1}^{2n}\tau_i.
            \end{equation}
        \AlAg{Individually, mask for hand along a ray $r$ can be obtained using Hand NeRF in Sec.~\ref{hand_nerf} as $M_{hand}(r)$. Similarly, masks for object (as shown in Fig.~\ref{fig:mask_homask}(b)) along a ray $r$ can be obtained using Object NeRF in Sec.~\ref{object_nerf} as $M_{obj}(r)$. Whereas, the background mask can be obtained using the foreground mask as $M_{bg}(r)=1-M_{fg}(r)$. Finally, along a ray $r$, the one-hot encoding of hand mask $M_{hand}(r)$, object mask $M_{obj}(r)$ and background mask $M_{bg}(r)$ gives us the predicted segmentation $M(r)$:
        \begin{equation}\label{eq:one-hot}
            M(r) = \arg max \{M_{hand}(r), M_{obj}(r), M_{bg}(r)\}.
        \end{equation}
        For stage 2, we also need visible object mask by following similar methodology as Eqs.~\ref{eq:nerfcolor},~\ref{eq:segmentation}. We can set the color prediction of the sample points for object to $1$, and color predictions for hand as $0$. Along a ray $r$, this allows only the rays from the visible object regions to transmit, and stops the transmittance from the object surface points occluded by hands, giving us visible object mask $M_{ho}$, as shown in Fig.~\ref{fig:mask_homask}(c).
        }
        \subsubsection{Pose Refinement}
            In this stage, we optimize both the hand and object pose in the camera-coordinate system, leveraging the composite rendering methodology. The initial hand and object poses are estimated. However, to achieve accurate alignment between the rendered and observed images, further optimization is performed to refine these pose estimates.
            The optimization process minimizes the discrepancy between the rendered and observed images using a differentiable renderer. Specifically, we iteratively adjust the hand and object poses, along with training compositional NeRF, by minimizing a loss function.
            RGB values of the predicted image are consistent with the ground truth image through the following loss function:
            \begin{equation}
                \mathcal{L}_{rgb1} = \Sigma_r||C(r) - \hat{C}(r)||,
            \end{equation}
            where $C(r)$ and $\hat{C}(r)$ are the predicted and groundtruth RGB colors along the ray $r$.
            Hand, object, and background masks are ensured consistent with ground truth by calculating the loss between predicted segmentation one-hot label $\AlAg{M}(r)$, calculated using Eq.~\ref{eq:one-hot}, and estimated ground truth \AlAg{segmentation obtained using SAM-Track~\cite{cheng2023segment} $\hat{M}(r)$, that takes image frames as input, and user input for target object in the image for the first frame, and outputs segmentation masks tracking the segmented object, following the same methodology as HOLD~\cite{fan2024hold}}:
            \begin{equation}
                \mathcal{L}_{seg1} = \Sigma_r||\AlAg{M}(r) - \AlAg{\hat{M}}(r)||.
            \end{equation}
            Eikonal loss $\mathcal{L}_{eikonal1}$ regularizes the canonical geometry, and $\mathcal{L}_{hand-sdf1}$ calculates the loss between predicted density and ground truth MANO model. Finally, a contact loss $\mathcal{L}_{contact1}$ encourages hand's fingertip vertices $V_{tip}$ to be close to object's vertices $V_o$, obtained via implicit networks:
            \begin{equation}
                \mathcal{L}_{contact1} = \Sigma_i \min_j ||V_{tip}^i - V_o^j||.
            \end{equation}
            The final loss that optimizes all the NeRF parameters, along with object and hand poses is defined as:
            \begin{equation}
            \begin{split}
                \mathcal{L}_{s1} = \mathcal{L}_{rgb1} + \lambda_{seg1}\mathcal{L}_{seg1} + \lambda_{eikonal1}\mathcal{L}_{eikonal1} + \\ \lambda_{hand-sdf1}\mathcal{L}_{hand-sdf1} + \lambda_{contact1}\mathcal{L}_{contact1},
            \end{split}
            \end{equation}
            where \AlAg{$\lambda_{seg1}$ is set to $1.1$ and progressively decreased to $0.1$ until $30,000$ iterations, $\lambda_{eikonal1}$ is $1$, $\lambda_{hand-sdf1}$ is set to $5.0$ and $\lambda_{contact1}$ is initially set to $0$ and progressively increased to $1.0$ until $30,000$ iterations.}
            \begin{figure}[ht]
                \centering
                \begin{tabular}{ccc}
                \includegraphics[width=0.3\linewidth]{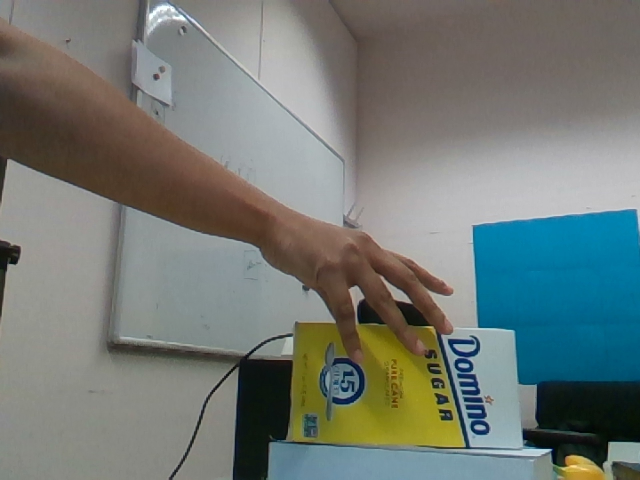} & \includegraphics[width=0.3\linewidth]{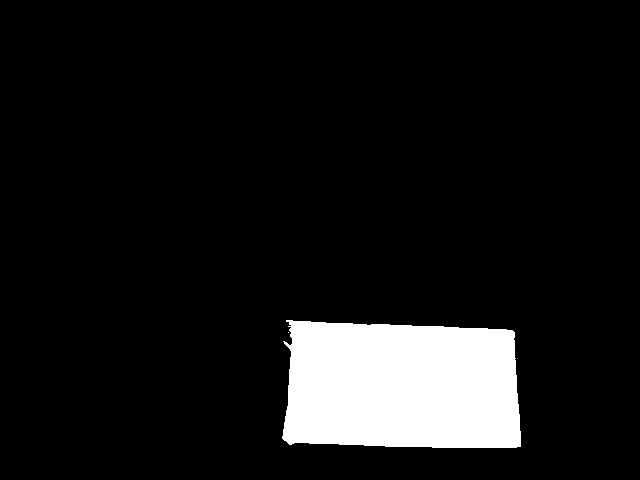} & \includegraphics[width=0.3\linewidth]{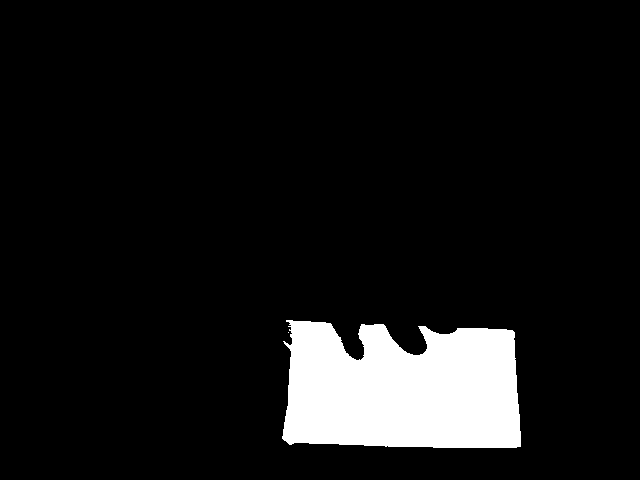} \\
                (a) & (b) & (c)
                \end{tabular}
                \caption{After Stage 1, for each (a) input image, predicted (b) object mask $M_{obj}$ and (c) hand-object mask $M_{ho}$ are calculated.}
                \label{fig:mask_homask}
            \end{figure}
            For a given frame (Fig.~\ref{fig:mask_homask} (a)), this stage prepares the object's implicit network, hand and object pose, object mask \AlAg{$M_{obj}$} (Fig.~\ref{fig:mask_homask} (b)) and hand-object mask \AlAg{$M_{ho}$} (Fig.~\ref{fig:mask_homask} (c)) for the second stage of our method.
    \subsection{Stage 2: Physics-based Inverse Rendering using Spherical Gaussians}
    \label{sec:stage2}
        In the second stage, we adopt a surface-based rendering approach to fine-tune the object’s texture. The surface rendering-based approach allows for optimizing surface-level material properties like roughness, reflectance, Fresnel, shadows, etc., as will be discussed in this section.
        
        This stage utilizes the refined masks and poses of the hand and object from Stage $1$, as well as the original image sequences captured from a monocular camera. The output includes high-fidelity geometry, albedo, specular maps, and hand visibility, in conjunction with environment illumination based on the object's learned material properties. By incorporating specularity and hand visibility, we prevent environmental factors such as shadows and reflections from being baked into the object’s texture, ensuring accurate albedo predictions. We will discuss the above-mentioned different elements of the Physics-Based Rendering (PBR) equation in this section.
        
        \subsubsection{Environment Illumination}
            \label{direct_illum}
            We implement environmental illumination using a sum of 128 Spherical Gaussians (SGs), which approximates the environmental light interacting with the object at any given point. These SGs represent light coming from various directions in the scene. This approximation allows us to efficiently simulate how the dynamic object is lit in an environment, ensuring that the lighting conditions are captured accurately in the final render.
            Each SG models the incident light distribution, and we apply these SGs to estimate the environmental contribution to the light reflected from the object's surface. This step enhances the realism by simulating how light behaves in real-world settings.
            Mathematically, from Eqs.~\eqref{eq:pbr} and ~\eqref{eq:sg}, the direct illumination can be represented as:
            \begin{equation}\label{eq:dir_illum}
                L_d(\omega_i) = \Sigma_{j=1}^{128}G(\omega_i;\psi_j,\mu_j,\AlAg{\eta}_j).
            \end{equation}
            Since the captured object is dynamic, in its object coordinate system the environment rotates in the counter direction of the object's transformation. Therefore, according to the input object pose, the lobe directions $\psi_j$ are rotated in the counter direction of the object's pose transform.
        \subsubsection{Bidirectional Radiance Distribution Field}
            Bidirectional Radiance Distribution Field (BRDF) on the surface point depends on diffuse albedo and specular effect. Diffuse BRDF is calculated using a Multi-layer Perceptron (MLP) predicting albedo color at the surface point $a(x)$, which defines the base color of the surface point $x$. Specular BRDF depends on material surface properties, like MLP-predicted roughness $r(x)$ and specular reflectance $s$. The roughness parameter controls how sharp or spread out the reflections are, and the specular reflectance parameter defines the metallic property of the object.
            Hence, the BRDF is the sum of diffuse BRDF $\phi_d$ and specular BRDF $\phi_s$:
            \begin{equation}
                \label{eq:brdf}
                \phi(x;\omega_o,\omega_i) = \phi_d(x) + \phi_s(x;\omega_o,\omega_i).
            \end{equation}
            The diffuse BRDF depends on predicted albedo color $a(x)$. The specular BRDF depends on Fresnel $F$ and self-shadowing $g$ terms, which are dependent on predicted specular reflectance $s$ and roughness $r(x)$.
            \begin{equation}
                \phi_d(x) = \frac{a(x)}{\pi}, \phi_s(x;\omega_o,\omega_i) = \frac{F(s,r(x))g(r(x))}{4(\omega_o \cdot n)(\omega_i \cdot n)},
            \end{equation}
            where $n$ is the surface normal on the point $x$.
        \subsubsection{Hand Occlusion}
            \label{hand_occ}
            One of the key challenges in rendering hand-object interactions is handling occlusion caused by the hand.
            To model this, we represent the hand as a set of $108$ parameterized spheres \AlAg{similar to power cells~\cite{wang2022computing}}. 
            We first place all spheres $\{p_i,r_i\}_{i=1}^{108}$ manually in the canonical MANO hand volume $\Psi$, with $p_i$ and $r_i$ being their centers and radii. 
            \begin{wrapfigure}{l}{0.25\textwidth}
                \centering
                \includegraphics[width=0.9\linewidth]{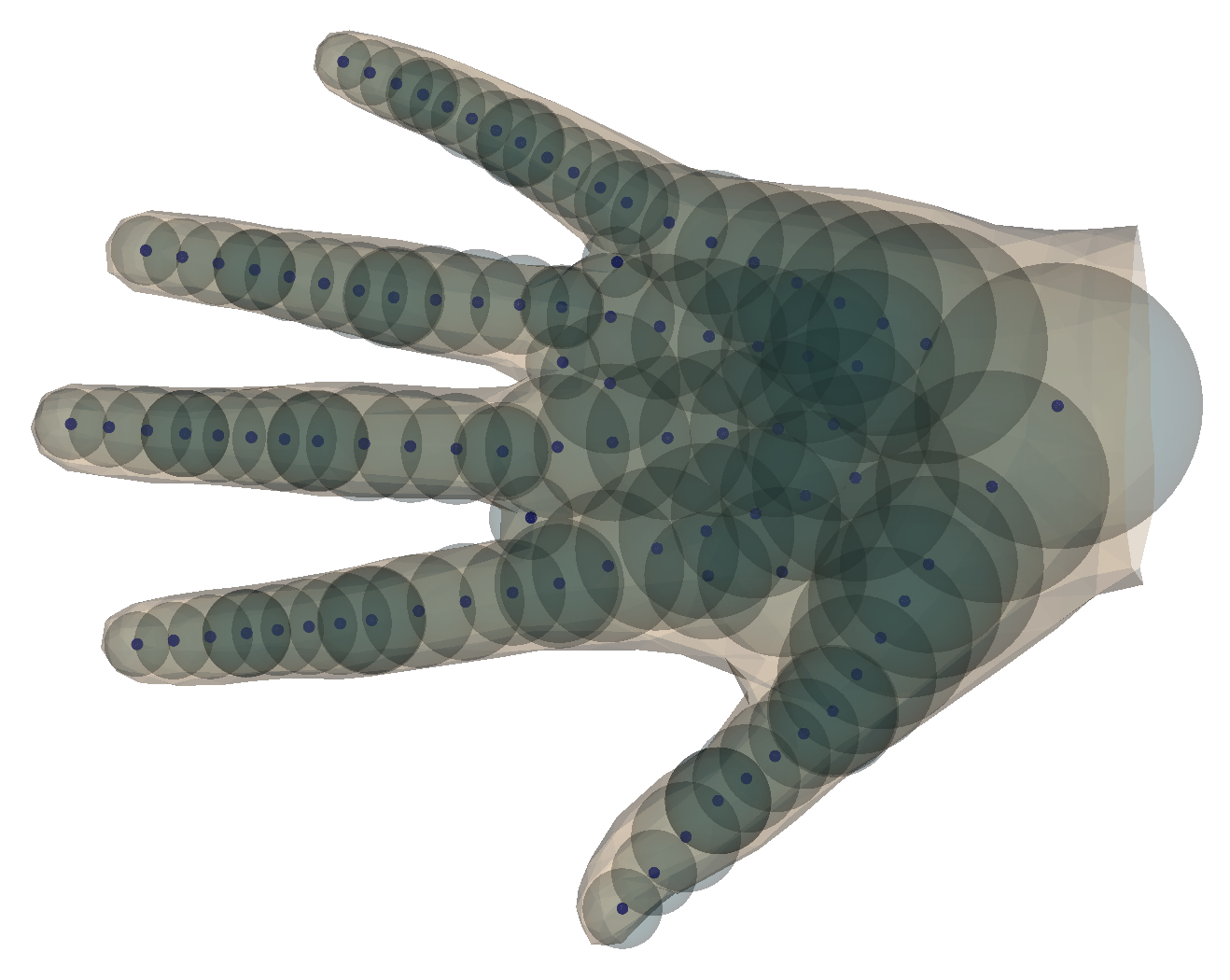}
                \caption{Canonical MANO hand is packed with $108$ parameterizable spheres for hand-occlusion computation.}
                \label{fig:mano-sphere}
            \end{wrapfigure}
            \AlAg{Initially, the sphere centers $\{p_i\}_{i=1}^{108}$ are determined on the canonical hand skeleton, manually selected between the MANO joints. The sphere radii $\{r_i\}_{i=1}^{108}$ requires partitioning the surface of the hand model into individual spheres.}
            These spheres define a power diagram that partitions the MANO hand volume $\Psi$ into a set of power cells.
            Each power cell $\Psi_{i}^{pow}$ consists of the points $x$ that are closest to a particular sphere $(c_i,r_i)$:
            \begin{equation}
                \Psi_{i}^{pow}: \{x\in\Psi | d_{pow}(x,c_i,r_i)\leq d_{pow}(x,c_j,r_j), j\neq i\},
            \end{equation}
            where $d_{pow}(x,p_i,r_i)=||x-p_i||^2-r_i^2$ is the power distance between the point $x$ and the sphere $(p_i,r_i)$. The power diagram partitions all vertices $\{v_j\}$ of canonical MANO hand model into each individual set $\mathcal{V}_i=\{v_j|v_j \in \Psi_{i}^{pow}\}$.
            Whenever the MANO hand model is deformed by its pose and shape parameters, the center $\hat{p}_i$ and radius $\hat{r}_i$ of each sphere can be updated correspondingly by the following rule: 
            \begin{equation}
                \begin{split}
                \hat{p}_i & = \frac{1}{|\mathcal{V}_i|} \sum_{v_j\in\mathcal{V}_i} \hat{v}_j, \\ 
                \hat{r}_i & = \frac{1}{|\mathcal{V}_i|} \sum_{v_j\in\mathcal{V}_i} || \hat{n}_j \cdot (\hat{v}_j - \hat{p}_i)||,
                \end{split}
            \end{equation}
            where $\hat{v}_j$ and $\hat{n}_j$ are the vertex position and normal associated with vertex $v_j$ in its deformed pose and shape.
            In this way, the centers and radii of these $108$  spheres are parameterized by the MANO hand shape and pose. 
            
\begin{figure}[ht]
                \centering
                \includegraphics[width=0.80\linewidth]{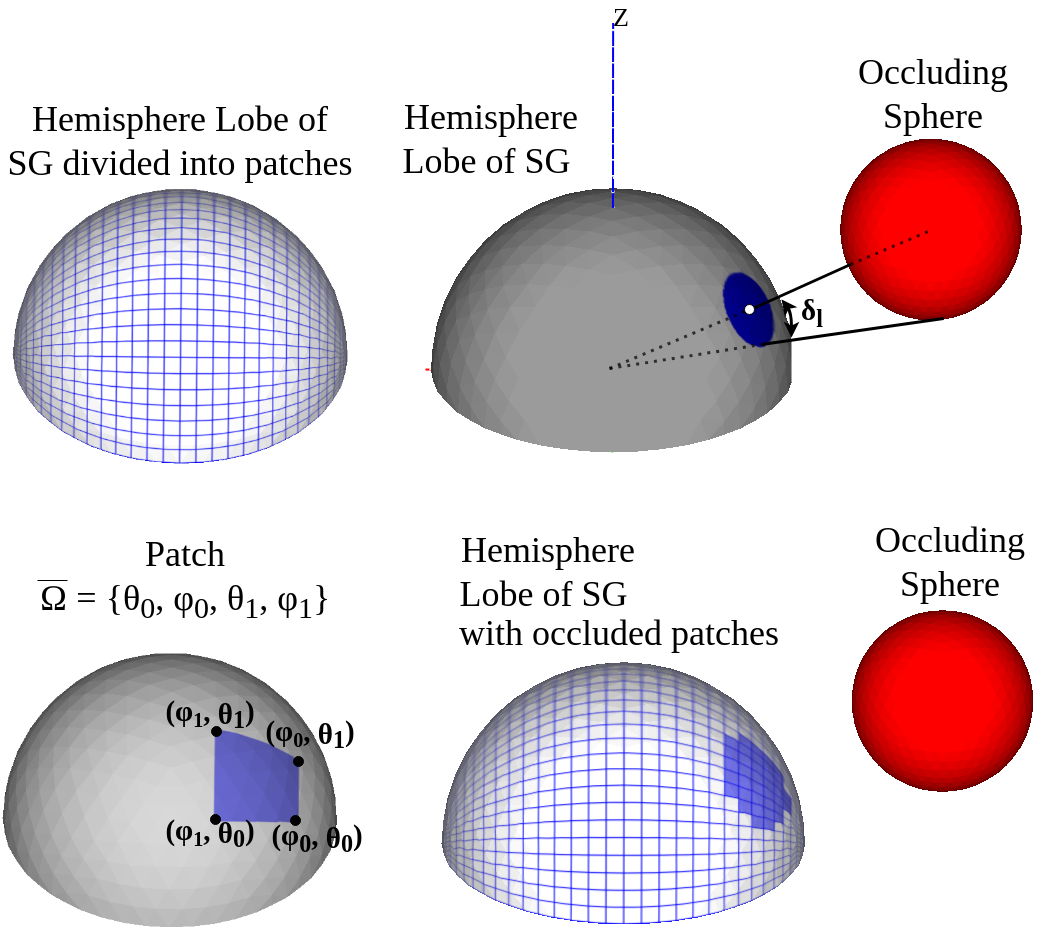}
                \caption{An SG hemispherical lobe, centered around the Z-axis, is divided into 64x64 patches. An occluding sphere projected onto the SG lobe covers some patches. The occluded patches are calculated, and their fractional value is calculated to get hand occlusion.}
                \label{fig:sphere-patch}
            \end{figure}

\begin{figure}[ht]
                \centering
                \includegraphics[width=0.8\linewidth]{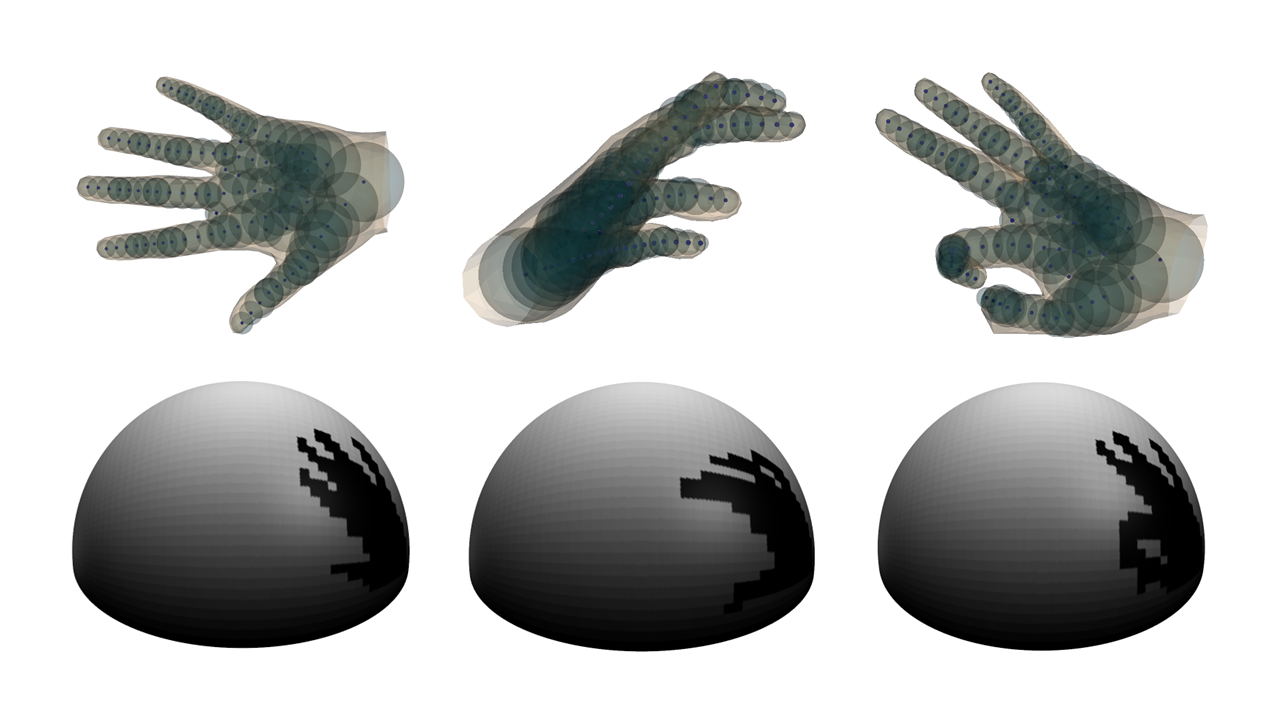}
                \caption{The parameterizable spheres are updated with hand pose transformation, and occlude the patches of Spherical Gaussian.}
                \label{fig:param-sphere}
            \end{figure}

            This granular representation allows us to model how different parts of the hand obstruct the light and cast shadows on the object's surface.
            For each surface point on the object, we divide its lobe of Spherical Gaussian (SG) into a grid of 64x64 patches, each representing a small region of the environment illumination and BRDF. Each one of the $108$ spheres representing MANO hand is projected onto the SG lobe, and the patches belonging to the occluded regions are collected. Finally, the fractional value of the integral of the occluded patches gives a measure of the extent to which each part of the hand blocks light from reaching the object. 
            For the following integral of a generic Spherical Gaussian (ISG) in polar format~\cite{zhang2021physg} with lobe sharpness \AlAg{$\eta$},
            \begin{equation}
                S(\theta, \phi, \AlAg{\eta}) = \int_0^\theta \int_0^\phi e^{\AlAg{\eta}(\cos \phi' - 1)} d\phi d\theta,
            \end{equation}
            where $\phi'$ is angle between the lobe direction of the Spherical Gaussian, and some direction $\omega_i$ along the hemisphere. $S(\theta, \phi, \AlAg{\eta})$ represents the accumulated Spherical Gaussian over a patch of the hemisphere, where $\theta$ and $\phi$ are angular coordinates. This means when $\theta=\pi$ and $\phi=\pi$, it covers the whole hemisphere.
            To compute the fractional value for a given angular region, we normalize by dividing the integral result by $S(\pi, \pi, \AlAg{\eta})$~\cite{iwasaki2012real}, as shown below:
            \begin{equation}
                \hat{S}(\theta, \phi, \AlAg{\eta}) = \frac{S(\theta, \phi, \AlAg{\eta})}{S(\pi, \pi, \AlAg{\eta})}.
            \end{equation}
            \begin{figure*}[ht]
                \centering
                \includegraphics[width=0.99\linewidth]{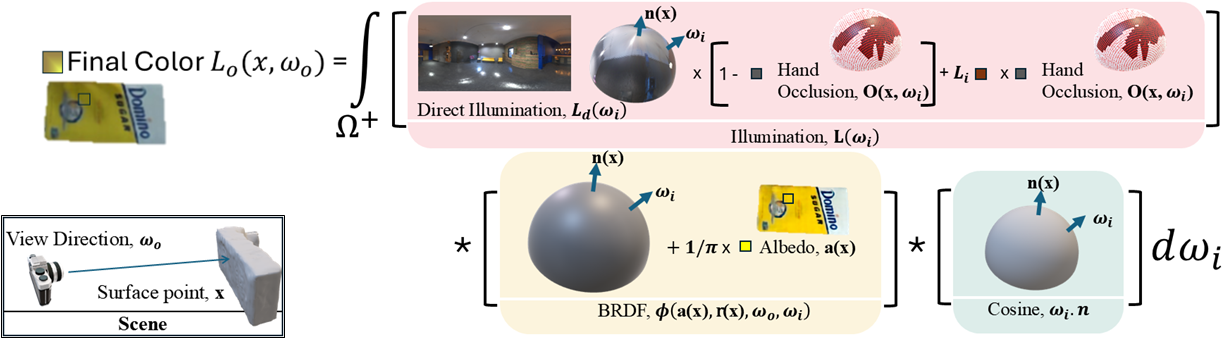}
                \caption{Physics-based rendering calculates an integral over a hemispherical region, centered around the surface normal of a surface point, to calculate the final color of the surface point. Based on SG approximation, direct illumination is defined as a sum of $128$ SGs, hand occlusion is calculated based on the parameterizable spherical representation of MANO hand, indirect illumination from occluding hand is a learned parameter, albedo represents the base color of the object without hand shadows or environment reflections, and specularity of the object is calculated based on the material properties - roughness, specular reflectance, etc.}
                \label{fig:pbr}
            \end{figure*}
            \AlAg{Iwasaki et. al.~\cite{iwasaki2012real} call this normalized term as normalized ISG, and approximate it as follows:
            \begin{equation} \label{eq:isg_approx}
                \hat{S}(\theta,\phi,\eta) \approx \frac{1}{1+e^{-g(\eta)(\theta-\pi/2)}}\frac{1}{1+e^{-h(\eta)(\phi-\pi/2)}},
            \end{equation}
            where they define $g$ and $h$ as fourth-degree polynomial function approximations over $\eta$ as follows:
            \begin{equation} \label{eq:isg_app_pol}
                \begin{split}
                    g(\eta) &= g_4(\eta/100)^4 + g_3(\eta/100)^3 + g_2(\eta/100)^2 + g_1(\eta/100)
                    \\
                    h(\eta) &= h_4(\eta/100)^4 + h_3(\eta/100)^3 + h_2(\eta/100)^2 + h_1(\eta/100),
                \end{split}
            \end{equation}
            where $g_4$, $g_3$, $g_2$, $g_1$, $h_4$, $h_3$, $h_2$, $h_1$ are constant values $-2.6856e^{-6}$, $7e^{-4}$,$-0.0571$, $3.9529$, $17.6028$, $-2.6875e^{-6}$, $7e^{-4}$, $-0.0592$, $3.9900$, $17.5003$ respectively.
            The patches $\Tilde{\Omega}$ occluded by hand spheres can be seen in Fig.~\ref{fig:param-sphere}.} The normalized value helps us calculate the proportion of the spherical Gaussian that falls within a specific patch of the sphere.
            \AlAg{From Eq.~\ref{eq:isg_approx} and Eq.~\ref{eq:isg_app_pol}}, for a patch defined \AlAg{along some direction $\omega_i$} by the angular coordinates $(\theta_0, \phi_0)$, $(\theta_1, \phi_0)$, $(\theta_0, \phi_1)$, and $(\theta_1, \phi_1)$ (See Fig.~\ref{fig:sphere-patch}), the final output is determined by computing the contributions from the \AlAg{patch corners}:
            \begin{equation}
                \begin{split}
                \AlAg{\mathcal{F}(x,\theta_0,\theta_1,\phi_0,\phi_1)} = \hat{S}(\theta_1, \phi_1, \AlAg{\eta}) - \hat{S}(\theta_1, \phi_0, \AlAg{\eta}) \\ - \hat{S}(\theta_0, \phi_1, \AlAg{\eta}) +
                \hat{S}(\theta_0, \phi_0, \AlAg{\eta}).   
                \end{split} 
            \end{equation}

            \AlAg{
            The sum over all the occluded patches gives us the fractional value $\mathcal{F}(x)$, 
            and is the fraction of ISG over occluded patches $\Tilde{\Omega}$.
            Therefore, we can approximate the integration of a partially-occluded Spherical Gaussian $G(\omega_i)O(x,\omega_i)$, with respect to the direction $\omega_i$ over the hemisphere $\Omega^+$ centered around the surface point $x$, using the above  integral over the occluded patches $\Tilde{\Omega}$ and its approximation $\mathcal{F}(x)$ as follows:
            \begin{equation} \label{eq:hand-occ}
                \begin{split}
                    \int_{\Omega^+} G(\omega_i)O(x,\omega_i)d\omega_i = \int_{\Tilde{\Omega}}G(\omega_i)d\omega_i \\ \approx \mathcal{F}(x)*\int_{\Omega^+} G(\omega_i)d\omega_i.
                \end{split}
            \end{equation}
            }
            
            
        \subsubsection{Final Illumination}
            We discussed direct illumination $L_d(\omega_i)$ from environment, as a sum of Spherical Gaussian along direction $w_i$ of hemisphere in \AlAg{Eq.~\ref{eq:dir_illum}}, and integral of SG with hand occlusion in \AlAg{Eq.~\ref{eq:hand-occ}}. In the regions of hand occlusion, the hand casts its own indirect illumination, which can be perceived as the hand's reflection on the object's surface. Hence, we learn a constant indirect illumination $L_i$, starting from a generic skin color ($\#E0AC69$) from an online search.
            The final illumination $L(\omega_i,x)$ from Eq.~\ref{eq:pbr} can thus be substituted as follows:
            \begin{equation}
                \label{eq:final_light}
                L(\omega_i,x) = L_d(\omega_i)*(1-O(x,\omega_i)) + L_i*O(x,\omega_i).
            \end{equation}
            Note that this equation also denotes a sum of $128$ SGs.
            To use the cosine term with other SG terms in PBR Eq.~\eqref{eq:pbr}, we approximate the cosine term $\omega_i \cdot n$, for incident direction $\omega_i$ at a surface point $x$ with surface normal $n$:
            \begin{equation}
                \label{eq:cosine}
                \omega_i \cdot n \approx G(\omega_i; 0.0315, 32.7080, n) - 31.7003.
            \end{equation}
            Combining illumination $L(\omega_i,x)$ from Eq.~\eqref{eq:final_light}, BRDF $\phi(x; \omega_o, \omega_i)$ from Eq.~\eqref{eq:brdf}, and cosine term $\omega_i \cdot n$ from Eq.~\eqref{eq:cosine}, we get the final color $c(\omega_o,x)$ at the surface point $x$ with view direction $\omega_o$ as the integral equation given in PBR Eq.~\eqref{eq:pbr}.
            To maintain the consistency of object shape, and avoid the hand's involvement in object geometry and texture reconstruction, we propose the following loss functions.
            An RGB loss is calculated on the color $c^{ho}$ predicted on $N_{ho}$ sampled pixels in the intersection of the predicted object mask and groundtruth hand-object mask \AlAg{$M_{ho}$ (Fig.~\ref{fig:mask_homask}(c))}:
            \begin{equation}
                \mathcal{L}_{rgb2} = \frac{1}{N_{ho}}\Sigma_{i=1}^{N_{ho}}||c_i^{ho} - \hat{c}_i||,
            \end{equation}
            where $\hat{c}_i$ is groundtruth color of the $N_{ho}$ sampled pixels.
            The below mask loss calculates minimum SDF $S^{no}$ on the $N_{no}$ rays corresponding to the sampled pixels outside the intersection of the predicted object mask and ground truth object mask \AlAg{$M_{obj}$ (Fig.~\ref{fig:mask_homask}(b))}:
            \begin{equation}
                \mathcal{L}_{mask2} = \frac{1}{N_{no}} \Sigma_{j=1}^{N_{no}} \frac{ln(1 + e^{-50*S_j^{no}})}{50}.
            \end{equation}
            Just like in stage 1, $\mathcal{L}_{eikonal2}$ regularizes object's geometry. The final loss optimized is as follows:
            \begin{equation}
                \mathcal{L}_{s2} = \mathcal{L}_{rgb2} + \lambda_{mask2}\mathcal{L}_{mask2} + \lambda_{eikonal2}\mathcal{L}_{eikonal2},
            \end{equation}
            \AlAg{where $\lambda_{mask2}$ is set to $100.0$ and $\lambda_{eikonal2}$ is $0.1$.}
    \subsection{Implementation Details}
        
        We utilize monocular camera capture datasets that encompass all viewpoints around the object, ensuring comprehensive coverage for high-fidelity texture and geometry learning. For the first stage, the 3D annotations estimated include hand-object pixel-wise segmentation~\cite{cheng2023segment,kirillov2023segment,yang2022deaot,yang2021aot,liu2023grounding,gong21b_interspeech}, hand MANO pose estimation~\cite{lin2021end}, and object pose estimation~\cite{sarlin2019coarse,sarlin2020superglue} in the camera-coordinate system. 
            
        The first stage of our pipeline involves the composite volumetric rendering of the hand, object, and background, following a pipeline similar to HOLD~\cite{fan2024hold}, using the initial estimates of the hand and object poses, along with segmentation masks. HOLD optimizes hand and object poses, geometry, and an initial estimate of texture.
        
        Using the optimized poses and object geometry from Stage 1, we derive the object mask and hand-object segmentation mask. These are then used in the second stage, which focuses on physics-based surface rendering of the object, following a pipeline inspired by PhySG~\cite{zhang2021physg}. A key modification in our method is the computation of occlusion to account for hand-object interaction, based on the approach in Iwasaki et al.~\cite{iwasaki2012real}. Prior works, such as PhySG, do not compute occlusion in this way, setting our method apart.
        In Stage 2, we optimize the following key parameters at a surface point $x$:
        \begin{itemize}
            \item Roughness \AlAg{$r(x)$} and albedo \AlAg{$a(x)$}, predicted using an MLP layer, which contribute to the specular BRDF \AlAg{$\phi_s(x)$} and diffuse BRDF \AlAg{$\phi_d(x)$} calculations, respectively.
            \item Spherical Gaussian (SG) parameters related to direct illumination \AlAg{$L_d(\omega_i)$} are optimized to accurately model light interactions with the object's surface.
            \item Additionally, an indirect illumination term \AlAg{$L_i$}, learned as an RGB value, accounts for the effects of hand shadows and reflections on the object.
        \end{itemize}
        It is important to note that Stage 2 is dependent on Stage 1, as it requires the hand-object pose and segmentation data obtained in the first stage. Therefore, Stage 2 is not object agnostic. However, Stage 1 is object agnostic, enabling the reconstruction of 3D geometry and texture for objects with unknown shapes and textures.
        Each scenario is trained on a 12 GB Nvidia GeForce RTX 2080 Ti GPU, and training takes approximately 14 hours for each stage.
        \begin{figure*}[ht]
            \centering
            \includegraphics[width=0.8\linewidth]{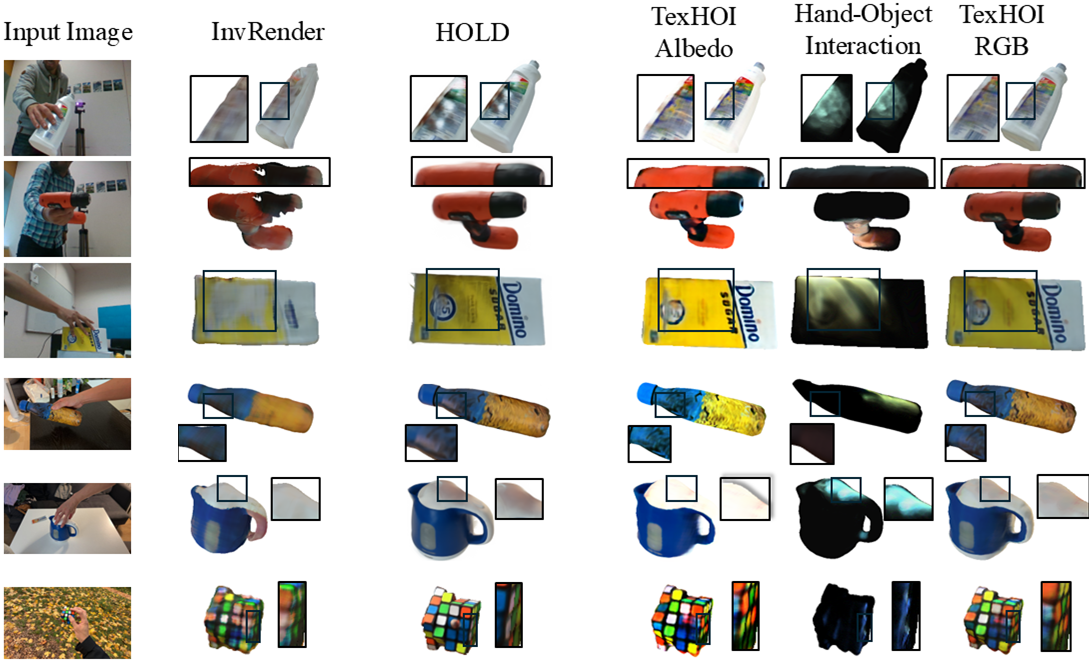}
            \caption{Qualitative Comparison: TexHOI albedo $a(x)$ predictions avoid baked hand-occlusions and environment-illumination. Hand occlusions can be observed in ``Hand-Object Interaction'' $\mathcal{F}(x)$ (0-1 normalized). The final TexHOI RGB $c(\omega_o, x)$ is calculated using Physics-Based Rendering, and is compared with other state-of-the-art approaches, HOLD~\cite{fan2024hold} and InvRender~\cite{zhang2022modeling}.}
            \label{fig:result}
        \end{figure*}
        \begin{table*}
            \centering
            \begin{tabular}{c|ccc|ccc|ccc}
                &  & \textbf{InvRender} & & & \textbf{HOLD} & & & \textbf{TexHOI} &\\
                \hline
                \textbf{HO3D Object} & PSNR $\uparrow$ & SSIM $\uparrow$ & LPIPS $\downarrow$ & PSNR $\uparrow$ & SSIM $\uparrow$ & LPIPS $\downarrow$ & PSNR $\uparrow$ & SSIM $\uparrow$ & LPIPS $\downarrow$ \\
                \hline
                Sugar Box           & 22.36 & 0.9501 & 0.110 & 22.20 & 0.9508 & 0.091 & \textbf{23.55} & \textbf{0.9543} & \textbf{0.067}\\
                Bleach Cleanser     & 18.76 & 0.9344 & 0.173 & 19.62 & 0.9383 & 0.109 & \textbf{20.78} & \textbf{0.9454} & \textbf{0.105}\\
                Drilling Machine    & 14.51 & 0.9097 & 0.222 & \textbf{18.12} & 0.9258 & 0.152 & 17.64 & \textbf{0.9259} & \textbf{0.139}\\
                Cracker Box         & 17.64 & 0.8910 & 0.197 & 19.52 & 0.8919 & 0.095 & \textbf{19.88} & \textbf{0.8972} & \textbf{0.090}\\
                Mustard Bottle      & 21.86 & 0.9476 & 0.138 & \textbf{22.08} & 0.9516 & \textbf{0.103} & 21.93 & \textbf{0.9550} & 0.110\\
                Meat Can            & 20.26 & 0.9491 & 0.085 & \textbf{22.16} & 0.9543 & \textbf{0.057} & 22.02 & \textbf{0.9547} & 0.060\\
                                        \hline
                \hline
                Average       & 19.23 & 0.9301 & 0.154 & 20.61 & 0.9354 & 0.101 & \textbf{20.96} & \textbf{0.9387} & \textbf{0.095}
            \end{tabular}
            \caption{Quantitative comparison is performed between our method, TexHOI, and other state-of-the-art approaches that reconstruct geometry and texture. HOLD bakes hand-occlusions and illumination on its texture prediction, InvRender does not handle dynamic occlusions.}
            \label{tab:result}
        \end{table*}
        \begin{figure}[ht]
            \centering
            \includegraphics[width=0.9\linewidth]{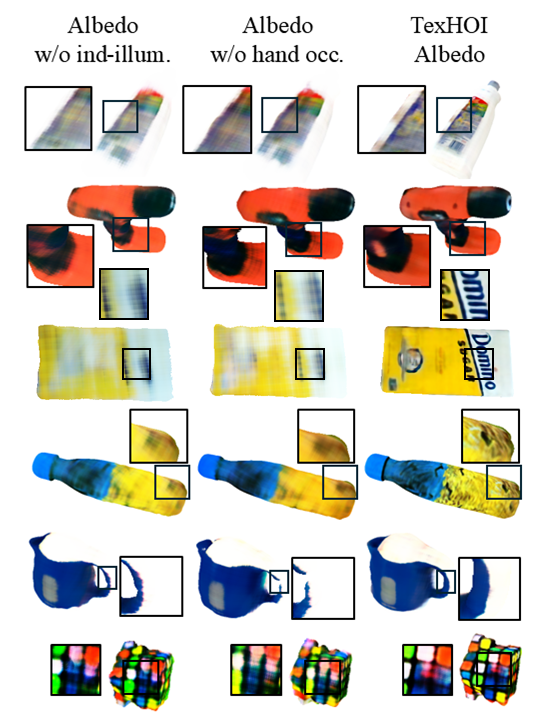}
            \caption{Ablation studies compare our model with the versions trained without indirect illumination, and without any occlusions. Illumination effects on surface albedo, and hand occlusions on surface can be observed. The scenes are in same order as Fig.~\ref{fig:result}}
            \label{fig:ablation}
        \end{figure}
        \begin{table*}[ht]
            \centering
            \begin{tabular}{c|ccc|ccc|ccc}
                &  & \textbf{w/o ind-illum} & & & \textbf{w/o occ} & & & \textbf{full} &\\
                \hline
                \textbf{HO3D Object} & PSNR $\uparrow$ & SSIM $\uparrow$ & LPIPS $\downarrow$ & PSNR $\uparrow$ & SSIM $\uparrow$ & LPIPS $\downarrow$ & PSNR $\uparrow$ & SSIM $\uparrow$ & LPIPS $\downarrow$ \\
                \hline
                Sugar Box           & 22.69 & \textbf{0.9548} & 0.113 & 23.54 & 0.9510 & 0.107 & \textbf{23.55} & 0.9543 & \textbf{0.067}\\
                Bleach Cleanser     & 20.45 & 0.9416 & 0.116 & 20.16 & 0.9382 & 0.118 & \textbf{20.78} & \textbf{0.9454} & \textbf{0.105}\\
                Drilling Machine    & 17.49 & \textbf{0.9263} & 0.146 & 17.58 & 0.9248 & 0.145 & \textbf{17.64} & 0.9259 & \textbf{0.139}\\
                Cracker Box         & 20.17 & 0.8971 & 0.124 & \textbf{20.22} & 0.8906 & 0.123 & 19.88 & \textbf{0.8972} & \textbf{0.090}\\
                Mustard Bottle      & 22.54 & 0.9544 & \textbf{0.108} & 21.77 & 0.9527 & 0.118 & \textbf{21.93} & \textbf{0.9550} & 0.110\\
                Meat Can            & 21.91 & 0.9541 & 0.067 & 21.83 & 0.9538 & 0.068 & \textbf{22.02} & \textbf{0.9547} & \textbf{0.060}\\
                \hline
                \hline
                Average           & 20.87 & 0.9380 & 0.112 & 20.85 & 0.9351 & 0.113 & \textbf{20.96} & \textbf{0.9387} & \textbf{0.095}
            \end{tabular}
            \caption{Quantitative ablative comparison of our model with the versions without indirect illumination, and without hand occlusions. Without illumination, the interacting hand casts a darker texture. Without occlusion handling, artifacts occur in major hand grasp regions.}
            \label{tab:ablation}
        \end{table*}
\section{Experiments}
        \subsection{Dataset}
            Our data consists of a sequence of the right hand of a subject interacting with a dynamic rigid object. To evaluate our approach, we use two types of datasets:
            \begin{figure}[ht]
                \centering
                \includegraphics[width=0.8\linewidth]{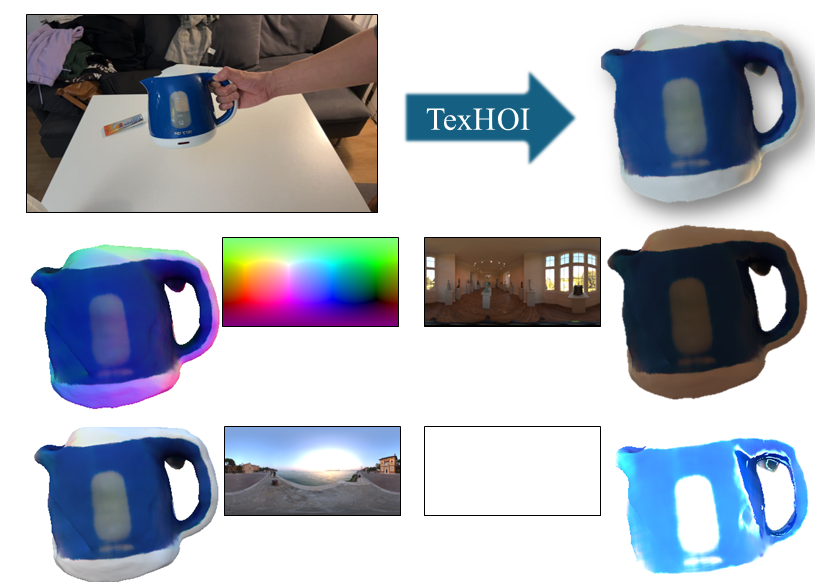}
                \caption{For an object reconstructed using our method, TexHOI, it can be relighted in different environments, based on different environment illumination.}
                \label{fig:relight}
            \end{figure}
            \begin{itemize}
                \item \textbf{HO3D}: The training data of HO3D~\cite{hampali2020honnotate} dataset contains 9 objects from YCB~\cite{xiang2017posecnn} models. We use monocular frames of six different objects from the dataset. These six objects were specifically selected based on the presence of sufficient texture details, which is essential for evaluating the effectiveness of our texture reconstruction. The six objects (and their subjects) are - sugar box (``ShSu''), cracker box (``MC''), bleach cleanser (``ABF''), meat can (``GPMF''), mustard bottle (``SM''), and drilling machine (``MDF''). The $3$ left objects in training data - scissors, banana, and mug contain no detailed-texture information for comparison.
                \item \textbf{In-the-wild}: We also use monocular frame sequences of three objects captured in real-world settings (``in-the-wild'')~\cite{fan2024hold}. These objects provide further diversity in testing by presenting challenging hand-object interactions and varied environmental conditions. The objects are - bottle, kettle, and rubrics cube.
            \end{itemize}
        \subsection{Baselines}
            We use HOLD~\cite{fan2024hold} as our primary baseline for performance comparison due to its proven effectiveness in dynamic hand-object interaction tasks, where it outperforms methods such as DiffHOI~\cite{ye2023diffusion} and BundleSDF~\cite{wen2023bundlesdf}. This makes HOLD an appropriate benchmark for assessing the improvements introduced by our method. 
            Our approach, however, goes beyond what HOLD achieves by focusing on texture reconstruction and illumination effects, areas that HOLD does not fully address. To highlight the improvements, we compare our full pipeline results with HOLD alone. This comparison demonstrates how our method enhances texture accuracy, particularly in capturing the complex interactions between hands and objects, such as shadows, reflections, and visibility changes, which are often overlooked in prior works.
            Additionally, for comparison with InvRender~\cite{zhang2022modeling}, we employ the 3D reconstruction from our Stage 1 and modify InvRender to incorporate object segmentation masks for computing pixel-wise losses in hand-object interaction images. We also adapt the approach by introducing Spherical Gaussian (SG) lobe rotations to effectively handle dynamic rigid object rotations, which are essential for handling both static and dynamic elements in the scene.
            \subsection{Evaluation Metrics}
            Since the objects manipulated in HO3D~\cite{hampali2020honnotate} dataset have their ground truth 3D models and textures in YCB~\cite{xiang2017posecnn}, we use them for our quantitative evaluation.
            For each video sequence in HO3D, we subsample one frame out of every 10 frames, resulting in around $200$-$400$ frames per sequence as our experimental data. We use Mitsuba~\cite{Mitsuba3} renderer and render the ground truth 3D models and textures to generate ground truth images corresponding to the poses of those sampled image frames, to compare our TexHOI's predicted albedo with other baseline methods. The quantitative evaluation metrics used are as follows:\\
            \begin{itemize}
                \item \textbf{Peak Signal-to-Noise Ratio (PSNR)}: This metric is used to evaluate the quality of the reconstructed texture by measuring how much signal (texture detail) is present compared to noise (artifacts or blurring).
                \item \textbf{Structural Similarity Index Measure (SSIM)}: We use SSIM to assess the visual similarity between our reconstructed textures and the ground truth. This metric helps to measure the perceptual quality of textures.
                \item \textbf{Learned Perceptual Image Patch Similarity (LPIPS)}: LPIPS is used to measure the perceptual distance between the predicted texture and the reference, focusing on perceptual similarity as seen by human vision. Lower LPIPS scores indicate higher perceptual quality.
            \end{itemize}
    \subsection{Baseline Comparisons}
        From Fig.~\ref{fig:result}, it is evident that while HOLD produces final images that closely resemble the input image, it achieves this by baking in the hand shadows and environmental reflections directly into the object’s texture. This, in turn, reduces the model's ability to generalize across different lighting environments and hand poses. In contrast, our method, TexHOI, can disentangle the object’s intrinsic texture (albedo) from external environmental influences, resulting in a cleaner and more accurate reconstruction of the object's surface.
        This observation is further supported by the evaluation metrics in Tab.~\ref{tab:result}. HOLD achieves better performance on metrics such as RGB similarity because it embeds hand shadows and environmental reflections into the predicted texture, making them appear closer to the ground truth in terms of pixel-wise accuracy. However, our method excels in producing albedo textures that are more accurate representations of the object's inherent properties, without conflating them with transient environmental elements.
    \subsection{Ablation Studies}
        To evaluate the importance of different components in our approach, we conduct ablation studies by removing specific features, such as indirect illumination and hand occlusion handling. Specifically, we compare the results of our full methodology against versions without indirect illumination modeling and without explicit hand occlusion consideration.
        Our results, from Fig.~\ref{fig:ablation} and Tab.~\ref{tab:ablation} indicate that indirect illumination significantly contributes to the realism of the reconstructed texture, particularly in complex scenes where light bounces off the occluding hand surface. Without this component, the reconstructed textures appear flatter and lack depth. Similarly, ignoring hand occlusions results in visual artifacts, as the shadowing and occlusion effects caused by interacting hands are not properly accounted for, leading to inconsistencies in the texture and geometry. These experiments highlight the necessity of modeling both indirect illumination and occlusions to achieve high-quality, realistic reconstructions. The few examples where the version without indirect illumination surpasses our full model in Tab.~\ref{tab:ablation} are again because the final prediction bakes the hand-environment interactions in the final prediction.
        \begin{figure}[ht]
            \centering
            \includegraphics[width=0.8\linewidth]{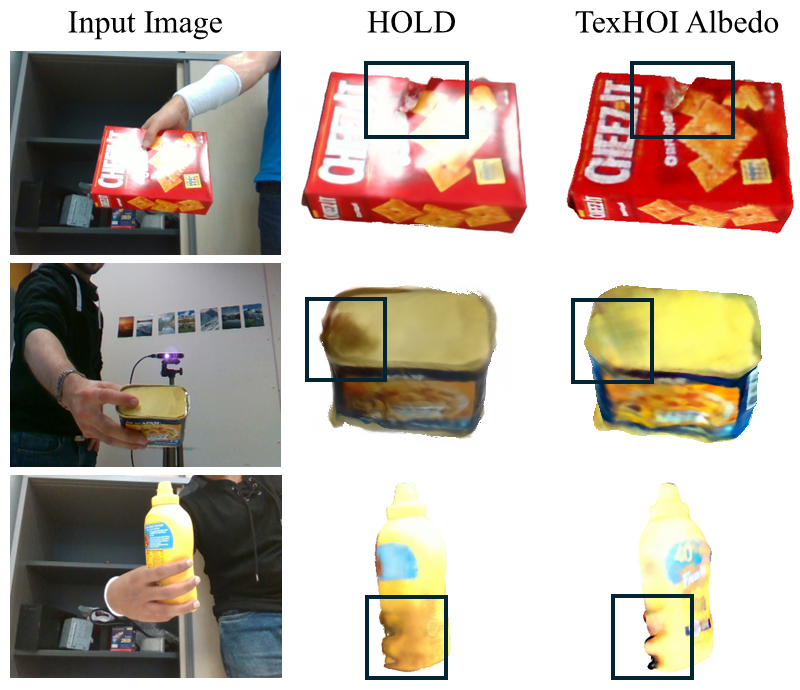}
            \caption{Our method is limited in geometry and texture reconstruction in cases when the hand is static/fixed w.r.t. object.}
            \label{fig:limitation}
        \end{figure}
    \subsection{Limitations}
        From Fig.~\ref{fig:limitation}, we observe that while our method is effective in generating high-detail geometry and textures, and successfully reconstructs albedo without baked-in environmental influences, it struggles in scenarios where the hand remains static relative to the object. In such cases, if the hand is not dynamic but remains fixed throughout the dataset, the impression of the hand tends to become baked into the geometry and texture of the object. This occurs because the model relies on changes in perspective and movement to distinguish between the object and interacting elements, and a lack of variability leads to ambiguity in the learned representation.
\section{Conclusion and Future Work}
    In this research, we demonstrate that TexHOI significantly outperforms existing state-of-the-art methods in reconstructing high-detail albedo textures free from environmental artifacts, particularly in challenging scenarios involving hand occlusions and environmental reflections. We achieve this by parameterizing occluding spheres, fitted into the MANO hand model, and calculating the fraction of occluded Spherical Gaussians to accurately represent hand occlusions.
    Furthermore, in many state-of-the-art methods, environmental illumination often reflects on the texture predictions, leading to inconsistencies when the object is viewed under different lighting conditions. Our approach tackles this problem by incorporating Physics-Based Rendering techniques, approximating illumination effects using Spherical Gaussians. This allows us to produce textures that are consistent across varying lighting environments, enhancing the robustness of the reconstructed model.
    
    Despite these advancements, there are several areas for future improvement. One potential direction is 
    incorporating a more sophisticated model for indirect illumination, particularly one that accounts for the specific properties of skin and other materials under different lighting conditions, could enhance the realism of the reconstructions.
    \AlAg{Although the centers of the hand spheres remain outside the object's surface, there can be inter-penetrations of the spheres with object near the fingertips. This causes negligible errors in the computation of hand occlusion, and can be an area of improvement in the future.}
    Another promising direction is exploring temporal consistency across frames in dynamic sequences, avoiding minor temporal artifacts. By incorporating temporal coherence into the reconstruction process, we could achieve smoother transitions and more stable textures in video sequences. 
    \AlAg{3D Gaussian splat-based inverse rendering have not been explored in the area of hand-object interactions, according to our knowledge, and can be a promising direction of progress in the research.}
    \AlAg{In the future, the work can also be optimized end-to-end using composite surface-based inverse rendering or a more efficient 3D Gaussian splat-based inverse rendering, eliminating the need for a 2-step process to correct poses and optimize texture, as all elements can be optimized simultaneously.}
    Finally, extending our model to handle a broad range of interaction scenarios, such as multiple hands, \AlAg{or handling inter-penetrations of hand and object using collision detection or physical simulation modules,} would make it more applicable to real-world use cases with complex interactions.


\begin{thebibliography}{10}\itemsep=-1pt
\bibliographystyle{IEEEtran}
\bibitem{Mitsuba3}
{Mitsuba3 Renderer}.

\bibitem{boss2021nerd}
M.~Boss, R.~Braun, V.~Jampani, J.~T. Barron, C.~Liu, and H.~Lensch.
\newblock {{NeRD}: Neural Reflectance Decomposition from Image Collections}.
\newblock In {\em International Conference on Computer Vision (ICCV)}, pages 12684--12694, 2021.

\bibitem{boss2022samurai}
M.~Boss, A.~Engelhardt, A.~Kar, Y.~Li, D.~Sun, J.~Barron, H.~Lensch, and V.~Jampani.
\newblock {{SAMURAI}: Shape and Material from Unconstrained Real-World Arbitrary Image Collections}.
\newblock {\em Advances in Neural Information Processing Systems (NeurIPS)}, 35:26389--26403, 2022.

\bibitem{boss2021neural}
M.~Boss, V.~Jampani, R.~Braun, C.~Liu, J.~Barron, and H.~Lensch.
\newblock {{Neural-PIL}: Neural Pre-Integrated Lighting for Reflectance Decomposition}.
\newblock {\em Advances in Neural Information Processing Systems (NeurIPS)}, 34:10691--10704, 2021.

\bibitem{brahmbhatt2020contactpose}
S.~Brahmbhatt, C.~Tang, C.~D. Twigg, C.~C. Kemp, and J.~Hays.
\newblock {{ContactPose}: A Dataset of Grasps with Object Contact and Hand Pose}.
\newblock In {\em European Conference on Computer Vision (ECCV)}, pages 361--378. Springer, 2020.

\bibitem{cai2024pbir}
G.~Cai, F.~Luan, M.~Ha{\v{s}}an, K.~Zhang, S.~Bi, Z.~Xu, I.~Georgiev, and S.~Zhao.
\newblock {{PBIR-NIE}: Glossy Object Capture under Non-Distant Lighting}.
\newblock {\em arXiv preprint arXiv:2408.06878}, 2024.

\bibitem{chao2021dexycb}
Y.-W. Chao, W.~Yang, Y.~Xiang, P.~Molchanov, A.~Handa, J.~Tremblay, Y.~S. Narang, K.~Van~Wyk, U.~Iqbal, S.~Birchfield, et~al.
\newblock {{DexYCB}: A Benchmark for Capturing Hand Grasping of Objects}.
\newblock In {\em Conference on Computer Vision and Pattern Recognition (CVPR)}, pages 9044--9053, 2021.

\bibitem{chen2022tensorf}
A.~Chen, Z.~Xu, A.~Geiger, J.~Yu, and H.~Su.
\newblock {{TensoRF}: Tensorial Radiance Fields}.
\newblock In {\em European Conference on Computer Vision (ECCV)}, pages 333--350. Springer, 2022.

\bibitem{chen2024intrinsicanything}
X.~Chen, S.~Peng, D.~Yang, Y.~Liu, B.~Pan, C.~Lv, and X.~Zhou.
\newblock {{IntrinsicAnything}: Learning Diffusion Priors for Inverse Rendering Under Unknown Illumination}.
\newblock {\em arXiv preprint arXiv:2404.11593}, 2024.

\bibitem{chen2022tracing}
Z.~Chen, C.~Ding, J.~Guo, D.~Wang, Y.~Li, X.~Xiao, W.~Wu, and L.~Song.
\newblock {{L-tracing}: Fast Light Visibility Estimation on Neural Surfaces by Sphere Tracing}.
\newblock In {\em European Conference on Computer Vision (ECCV)}, pages 217--233. Springer, 2022.

\bibitem{cheng2023segment}
Y.~Cheng, L.~Li, Y.~Xu, X.~Li, Z.~Yang, W.~Wang, and Y.~Yang.
\newblock {Segment and Track Anything}.
\newblock {\em arXiv preprint arXiv:2305.06558}, 2023.

\bibitem{choi2024handnerf}
H.~Choi, N.~Chavan-Dafle, J.~Yuan, V.~Isler, and H.~Park.
\newblock {{HandNeRF}: Learning to Reconstruct Hand-Object Interaction Scene from a Single RGB Image}.
\newblock In {\em International Conference on Robotics and Automation (ICRA)}, pages 13940--13946. IEEE, 2024.

\bibitem{curless1996volumetric}
B.~Curless and M.~Levoy.
\newblock {A Volumetric Method for Building Complex Models from Range Images}.
\newblock In {\em Conference on Computer Graphics and Interactive Techniques (SIGGRAPH)}, pages 303--312, 1996.

\bibitem{dai2024mirres}
Y.~Dai, Q.~Wang, J.~Zhu, D.~Xi, Y.~Huo, C.~Qian, and Y.~He.
\newblock {{MIRReS}: Multi-bounce Inverse Rendering using Reservoir Sampling}.
\newblock {\em arXiv preprint arXiv:2406.16360}, 2024.

\bibitem{driess2023learning}
D.~Driess, Z.~Huang, Y.~Li, R.~Tedrake, and M.~Toussaint.
\newblock {Learning Multi-Object Dynamics with Compositional Neural Radiance Fields}.
\newblock In {\em Conference on Robot Learning (CoRL)}, pages 1755--1768. PMLR, 2023.

\bibitem{eck1995multiresolution}
M.~Eck, T.~DeRose, T.~Duchamp, H.~Hoppe, M.~Lounsbery, and W.~Stuetzle.
\newblock {Multiresolution Analysis of Arbitrary Meshes}.
\newblock In {\em Conference on Computer Graphics and Interactive Techniques (SIGGRAPH)}, pages 173--182, 1995.

\bibitem{fan2024hold}
Z.~Fan, M.~Parelli, M.~E. Kadoglou, X.~Chen, M.~Kocabas, M.~J. Black, and O.~Hilliges.
\newblock {{HOLD}: Category-agnostic 3D Reconstruction of Interacting Hands and Objects from Video}.
\newblock In {\em Conference on Computer Vision and Pattern Recognition (CVPR)}, pages 494--504, 2024.

\bibitem{gao2023relightable}
J.~Gao, C.~Gu, Y.~Lin, H.~Zhu, X.~Cao, L.~Zhang, and Y.~Yao.
\newblock {{Relightable 3D Gaussian}: Real-Time Point Cloud Relighting with BRDF Decomposition and Ray Tracing}.
\newblock {\em arXiv preprint arXiv:2311.16043}, 2023.

\bibitem{gong21b_interspeech}
Y.~Gong, Y.-A. Chung, and J.~Glass.
\newblock {{AST}: Audio Spectrogram Transformer}.
\newblock In {\em Proc. Interspeech}, pages 571--575, 2021.

\bibitem{guo2023handnerf}
Z.~Guo, W.~Zhou, M.~Wang, L.~Li, and H.~Li.
\newblock {{HandNeRF}: Neural Radiance Fields for Animatable Interacting Hands}.
\newblock In {\em Conference on Computer Vision and Pattern Recognition (CVPR)}, pages 21078--21087, 2023.

\bibitem{hampali2020honnotate}
S.~Hampali, M.~Rad, M.~Oberweger, and V.~Lepetit.
\newblock {{HOnnotate}: A Method for 3D Annotation of Hand and Object Poses}.
\newblock In {\em Conference on Computer Vision and Pattern Recognition}, pages 3196--3206, 2020.

\bibitem{hasselgren2022shape}
J.~Hasselgren, N.~Hofmann, and J.~Munkberg.
\newblock {Shape, Light, and Material Decomposition from Images using Monte Carlo Rendering and Denoising}.
\newblock {\em Advances in Neural Information Processing Systems (NeurIPS)}, 35:22856--22869, 2022.

\bibitem{iwasaki2012real}
K.~Iwasaki, W.~Furuya, Y.~Dobashi, and T.~Nishita.
\newblock {Real-time Rendering of Dynamic Scenes under All-Frequency Lighting using Integral Spherical Gaussian}.
\newblock In {\em Computer Graphics Forum}, volume~31, pages 727--734. Wiley Online Library, 2012.

\bibitem{james2012straightforward}
M.~R. James and S.~Robson.
\newblock {Straightforward Reconstruction of 3D Surfaces and Topography with a Camera: Accuracy and Geoscience Application}.
\newblock {\em Journal of Geophysical Research: Earth Surface}, 117(F3), 2012.

\bibitem{jin2023tensoir}
H.~Jin, I.~Liu, P.~Xu, X.~Zhang, S.~Han, S.~Bi, X.~Zhou, Z.~Xu, and H.~Su.
\newblock {{TensoIR}: Tensorial Inverse Rendering}.
\newblock In {\em Conference on Computer Vision and Pattern Recognition (CVPR)}, pages 165--174, 2023.

\bibitem{kawai1993radioptimization}
J.~K. Kawai, J.~S. Painter, and M.~F. Cohen.
\newblock {Radioptimization: Goal based Rendering}.
\newblock In {\em Conference on Computer Graphics and Interactive Techniques (SIGGRAPH)}, pages 147--154, 1993.

\bibitem{kerbl20233d}
B.~Kerbl, G.~Kopanas, T.~Leimk{\"u}hler, and G.~Drettakis.
\newblock {3D Gaussian Splatting for Real-Time Radiance Field Rendering}.
\newblock {\em ACM Transactions on Graphics (ToG)}, 42(4):139--1, 2023.

\bibitem{kirillov2023segment}
A.~Kirillov, E.~Mintun, N.~Ravi, H.~Mao, C.~Rolland, L.~Gustafson, T.~Xiao, S.~Whitehead, A.~C. Berg, W.-Y. Lo, et~al.
\newblock {Segment Anything}.
\newblock {\em arXiv preprint arXiv:2304.02643}, 2023.

\bibitem{knodt2021neural}
J.~Knodt, J.~Bartusek, S.-H. Baek, and F.~Heide.
\newblock {{Neural Ray-Tracing}: Learning Surfaces and Reflectance for Relighting and View Synthesis}.
\newblock {\em arXiv preprint arXiv:2104.13562}, 2021.

\bibitem{li2024tensosdf}
J.~Li, L.~Wang, L.~Zhang, and B.~Wang.
\newblock {{TensoSDF}: Roughness-aware Tensorial Representation for Robust Geometry and Material Reconstruction}.
\newblock {\em arXiv preprint arXiv:2402.02771}, 2024.

\bibitem{lin2021end}
K.~Lin, L.~Wang, and Z.~Liu.
\newblock {End-to-End Human Pose and Mesh Reconstruction with Transformers}.
\newblock In {\em Conference on Computer Vision and Pattern Recognition (CVPR)}, pages 1954--1963, 2021.

\bibitem{liu2023grounding}
S.~Liu, Z.~Zeng, T.~Ren, F.~Li, H.~Zhang, J.~Yang, C.~Li, J.~Yang, H.~Su, J.~Zhu, et~al.
\newblock {{Grounding Dino}: Marrying Dino with Grounded Pre-Training for Open-Set Object Detection}.
\newblock {\em arXiv preprint arXiv:2303.05499}, 2023.

\bibitem{liu2023nero}
Y.~Liu, P.~Wang, C.~Lin, X.~Long, J.~Wang, L.~Liu, T.~Komura, and W.~Wang.
\newblock {{NeRO}: Neural Geometry and BRDF Reconstruction of Reflective Objects from Multiview Images}.
\newblock {\em ACM Transactions on Graphics (ToG)}, 42(4):1--22, 2023.

\bibitem{marschner1998inverse}
S.~R. Marschner.
\newblock {\em {Inverse Rendering for Computer Graphics}}.
\newblock Cornell University, 1998.

\bibitem{mildenhall2021nerf}
B.~Mildenhall, P.~P. Srinivasan, M.~Tancik, J.~T. Barron, R.~Ramamoorthi, and R.~Ng.
\newblock {{NeRF}: Representing Scenes as Neural Radiance Fields for View Synthesis}.
\newblock {\em Communications of the ACM}, 65(1):99--106, 2021.

\bibitem{muller2022instant}
T.~M{\"u}ller, A.~Evans, C.~Schied, and A.~Keller.
\newblock {Instant Neural Graphics Primitives with a Multiresolution Hash Encoding}.
\newblock {\em ACM Transactions on Graphics (ToG)}, 41(4):1--15, 2022.

\bibitem{nimeroff1995efficient}
J.~S. Nimeroff, E.~Simoncelli, and J.~Dorsey.
\newblock {Efficient Re-rendering of Naturally Illuminated Environments}.
\newblock In {\em Photorealistic Rendering Techniques}, pages 373--388. Springer, 1995.

\bibitem{ozyecsil2017survey}
O.~{\"O}zye{\c{s}}il, V.~Voroninski, R.~Basri, and A.~Singer.
\newblock {A Survey of Structure from Motion*.}
\newblock {\em Acta Numerica}, 26:305--364, 2017.

\bibitem{romero2022embodied}
J.~Romero, D.~Tzionas, and M.~J. Black.
\newblock {{Embodied Hands}: Modeling and Capturing Hands and Bodies Together}.
\newblock {\em arXiv preprint arXiv:2201.02610}, 2022.

\bibitem{sarlin2019coarse}
P.-E. Sarlin, C.~Cadena, R.~Siegwart, and M.~Dymczyk.
\newblock {From Coarse to Fine: Robust Hierarchical Localization at Large Scale}.
\newblock In {\em Conference on Computer Vision and Pattern Recognition (CVPR)}, 2019.

\bibitem{sarlin2020superglue}
P.-E. Sarlin, D.~DeTone, T.~Malisiewicz, and A.~Rabinovich.
\newblock {{SuperGlue}: Learning Feature Matching with Graph Neural Networks}.
\newblock In {\em Conference on Computer Vision and Pattern Recognition (CVPR)}, 2020.

\bibitem{srinivasan2021nerv}
P.~P. Srinivasan, B.~Deng, X.~Zhang, M.~Tancik, B.~Mildenhall, and J.~T. Barron.
\newblock {{NeRV}: Neural Reflectance and Visibility Fields for Relighting and View Synthesis}.
\newblock In {\em Conference on Computer Vision and Pattern Recognition (CVPR)}, pages 7495--7504, 2021.

\bibitem{su2014estimating}
H.~Su, Q.~Huang, N.~J. Mitra, Y.~Li, and L.~Guibas.
\newblock {Estimating Image Depth using Shape Collections}.
\newblock {\em ACM Transactions on Graphics (ToG)}, 33(4):1--11, 2014.

\bibitem{sun2023neural}
C.~Sun, G.~Cai, Z.~Li, K.~Yan, C.~Zhang, C.~Marshall, J.-B. Huang, S.~Zhao, and Z.~Dong.
\newblock {Neural-PBIR Reconstruction of Shape, Material, and Illumination}.
\newblock In {\em International Conference on Computer Vision (ICCV)}, pages 18046--18056, 2023.

\bibitem{sun2022improved}
C.~Sun, M.~Sun, and H.-T. Chen.
\newblock {Improved Direct Voxel Grid Optimization for Radiance Fields Reconstruction}.
\newblock {\em arXiv preprint arXiv:2206.05085}, 2022.

\bibitem{teo1997cient}
P.~C. Teo, E.~P. Simoncelli, and D.~J. Heeger.
\newblock {Efficient Linear Re-rendering for Interactive Lighting Design}.
\newblock Technical report, Citeseer, 1997.

\bibitem{turk1991generating}
G.~Turk.
\newblock {Generating Textures on Arbitrary Surfaces using Reaction-Diffusion}.
\newblock {\em Conference on Computer Graphics and Interactive Techniques (SIGGRAPH)}, 25(4):289--298, 1991.

\bibitem{refnerf}
D.~Verbin, P.~Hedman, B.~Mildenhall, T.~Zickler, J.~T. Barron, and P.~P. Srinivasan.
\newblock {{Ref-NeRF}: Structured View-Dependent Appearance for Neural Radiance Fields}.
\newblock {\em Computer Vision and Pattern Recognition (CVPR)}, 2022.

\bibitem{wang2024inverse}
H.~Wang, W.~Hu, L.~Zhu, and R.~W. Lau.
\newblock {Inverse Rendering of Glossy Objects via the Neural Plenoptic Function and Radiance Fields}.
\newblock In {\em Conference on Computer Vision and Pattern Recognition (CVPR)}, pages 19999--20008, 2024.

\bibitem{wang2022computing}
N.~Wang, B.~Wang, W.~Wang, and X.~Guo.
\newblock Computing medial axis transform with feature preservation via restricted power diagram.
\newblock {\em ACM Transactions on Graphics (TOG)}, 41(6):1--18, 2022.

\bibitem{wang2021neus}
P.~Wang, L.~Liu, Y.~Liu, C.~Theobalt, T.~Komura, and W.~Wang.
\newblock {{NeuS}: Learning Neural Implicit Surfaces by Volume Rendering for Multi-View Reconstruction}.
\newblock {\em arXiv preprint arXiv:2106.10689}, 2021.

\bibitem{wang2023learning}
Y.~Wang, W.~Wu, and D.~Xu.
\newblock {Learning Unified Decompositional and Compositional {NeRF} for Editable Novel View Synthesis}.
\newblock In {\em Conference on Computer Vision and Pattern Recognition (CVPR)}, pages 18247--18256, 2023.

\bibitem{wen2023bundlesdf}
B.~Wen, J.~Tremblay, V.~Blukis, S.~Tyree, T.~M{\"u}ller, A.~Evans, D.~Fox, J.~Kautz, and S.~Birchfield.
\newblock {{BundleSDF}: Neural 6-DoF Tracking and 3D Reconstruction of Unknown Objects}.
\newblock In {\em Conference on Computer Vision and Pattern Recognition (CVPR)}, pages 606--617, 2023.

\bibitem{westoby2012structure}
M.~J. Westoby, J.~Brasington, N.~F. Glasser, M.~J. Hambrey, and J.~M. Reynolds.
\newblock {{`Structure-from-Motion' photogrammetry}: A low-cost, effective tool for geoscience applications}.
\newblock {\em Geomorphology}, 179:300--314, 2012.

\bibitem{witkin1991reaction}
A.~Witkin and M.~Kass.
\newblock {Reaction-Diffusion Textures}.
\newblock In {\em Conference on Computer Graphics and Interactive Techniques (SIGGRAPH)}, pages 299--308, 1991.

\bibitem{wu2022object}
Q.~Wu, X.~Liu, Y.~Chen, K.~Li, C.~Zheng, J.~Cai, and J.~Zheng.
\newblock {Object-Compositional Neural Implicit Surfaces}.
\newblock In {\em European Conference on Computer Vision (ECCV)}, pages 197--213. Springer, 2022.

\bibitem{xiang2017posecnn}
Y.~Xiang, T.~Schmidt, V.~Narayanan, and D.~Fox.
\newblock {PoseCNN}: A convolutional neural network for 6d object pose estimation in cluttered scenes.
\newblock {\em arXiv preprint arXiv:1711.00199}, 2017.

\bibitem{yan2023nerf}
Z.~Yan, C.~Li, and G.~H. Lee.
\newblock {{NeRF-DS}: Neural Radiance Fields for Dynamic Specular Objects}.
\newblock In {\em Conference on Computer Vision and Pattern Recognition (CVPR)}, pages 8285--8295, 2023.

\bibitem{yang2021learning}
B.~Yang, Y.~Zhang, Y.~Xu, Y.~Li, H.~Zhou, H.~Bao, G.~Zhang, and Z.~Cui.
\newblock {Learning Object-Compositional Neural Radiance Field for Editable Scene Rendering}.
\newblock In {\em International Conference on Computer Vision (ICCV)}, pages 13779--13788, 2021.

\bibitem{yang2022ps}
W.~Yang, G.~Chen, C.~Chen, Z.~Chen, and K.-Y.~K. Wong.
\newblock {{PS-NeRF}: Neural Inverse Rendering for Multi-View Photometric Stereo}.
\newblock In {\em European Conference on Computer Vision (ECCV)}, pages 266--284. Springer, 2022.

\bibitem{yang2021aot}
Z.~Yang, Y.~Wei, and Y.~Yang.
\newblock {Associating Objects with Transformers for Video Object Segmentation}.
\newblock In {\em Advances in Neural Information Processing Systems (NeurIPS)}, 2021.

\bibitem{yang2022deaot}
Z.~Yang and Y.~Yang.
\newblock {Decoupling Features in Hierarchical Propagation for Video Object Segmentation}.
\newblock In {\em Advances in Neural Information Processing Systems (NeurIPS)}, 2022.

\bibitem{yao2022neilf}
Y.~Yao, J.~Zhang, J.~Liu, Y.~Qu, T.~Fang, D.~McKinnon, Y.~Tsin, and L.~Quan.
\newblock {{NeILF}: Neural Incident Light Field for Physically-based Material Estimation}.
\newblock In {\em European Conference on Computer Vision (ECCV)}, pages 700--716. Springer, 2022.

\bibitem{yariv2020multiview}
L.~Yariv, Y.~Kasten, D.~Moran, M.~Galun, M.~Atzmon, B.~Ronen, and Y.~Lipman.
\newblock {Multiview Neural Surface Reconstruction by Disentangling Geometry and Appearance}.
\newblock {\em Advances in Neural Information Processing Systems (NeurIPS)}, 33:2492--2502, 2020.

\bibitem{ye2023diffusion}
Y.~Ye, P.~Hebbar, A.~Gupta, and S.~Tulsiani.
\newblock {Diffusion-Guided Reconstruction of Everyday Hand-Object Interaction }.
\newblock In {\em International Conference on Computer Vision (ICCV)}, pages 19717--19728, 2023.

\bibitem{zhang2022iron}
K.~Zhang, F.~Luan, Z.~Li, and N.~Snavely.
\newblock {{IRON}: Inverse Rendering by Optimizing Neural SDFs and Materials from Photometric Images}.
\newblock In {\em Conference on Computer Vision and Pattern Recognition (CVPR)}, pages 5565--5574, 2022.

\bibitem{zhang2021physg}
K.~Zhang, F.~Luan, Q.~Wang, K.~Bala, and N.~Snavely.
\newblock {{PhySG}: Inverse Rendering with Spherical Gaussians for Physics-based Material Editing and Relighting}.
\newblock In {\em Conference on Computer Vision and Pattern Recognition (CVPR)}, pages 5453--5462, 2021.

\bibitem{zhang2021nerfactor}
X.~Zhang, P.~P. Srinivasan, B.~Deng, P.~Debevec, W.~T. Freeman, and J.~T. Barron.
\newblock {{NeRFactor}: Neural Factorization of Shape and Reflectance under an Unknown Illumination}.
\newblock {\em ACM Transactions on Graphics (ToG)}, 40(6):1--18, 2021.

\bibitem{zhang2022modeling}
Y.~Zhang, J.~Sun, X.~He, H.~Fu, R.~Jia, and X.~Zhou.
\newblock {Modeling Indirect Illumination for Inverse Rendering}.
\newblock In {\em Conference on Computer Vision and Pattern Recognition (CVPR)}, pages 18643--18652, 2022.

\end{thebibliography}
\end{document}